\documentclass[sigconf]{acmart}

\setcopyright{acmcopyright}
\copyrightyear{2022}
\acmYear{2022}
\acmDOI{3514094.3534150}

\acmConference[AIES '22]{August 01--03, 2022}{Oxford, UK}

\usepackage{array}
\newcolumntype{M}[1]{>{\centering\arraybackslash}m{#1}}

\usepackage[super]{nth}

\usepackage{pdfpages}

\usepackage{xcolor}

\begin{document}

\title{Do Humans Trust Advice More if it Comes from AI? An Analysis of Human-AI Interactions}

\author{Kailas Vodrahalli}
\email{kailasv@stanford.edu}
\affiliation{%
  \institution{Stanford University}
  \country{Stanford, USA}
}

\author{Roxana Daneshjou}
\email{roxanad@stanford.edu}
\affiliation{%
  \institution{Stanford University}
  \country{Stanford, USA}
}

\author{Tobias Gerstenberg}
\email{gerstenberg@stanford.edu}
\affiliation{%
  \institution{Stanford University}
  \country{Stanford, USA}
}

\author{James Zou}
\email{jamesz@stanford.edu}
\affiliation{%
  \institution{Stanford University}
  \country{Stanford, USA}
}

\renewcommand{\shortauthors}{Vodrahalli, et al.}

\begin{abstract}
In decision support applications of AI, the AI algorithm's output is framed as a suggestion to a human user. The user may ignore this advice or take it into consideration to modify their decision. 
With the increasing prevalence of such human-AI interactions, it is important to understand how users react to AI advice.    
In this paper, we recruited over 1100 crowdworkers to characterize how humans use AI suggestions relative to equivalent suggestions from a group of peer humans across several experimental settings. We find that participants' beliefs about how human versus AI performance on a given task affects whether they heed the advice. 
When participants do heed the advice, they use it similarly for human and AI suggestions. Based on these results, we propose a two-stage, ``activation-integration'' model for human behavior and use it to characterize the factors that affect human-AI interactions.
\end{abstract}

\ccsdesc[500]{Human-centered computing~User models}
\ccsdesc[500]{Human-centered computing~Empirical studies in HCI}
\ccsdesc[500]{Human-centered computing~HCI theory, concepts and models}

\keywords{artificial intelligence, AI advice, human-in-the-loop, human interaction with AI, AI trust}

\maketitle

\section{Introduction}
In safety-critical areas like medicine, there remain significant barriers to widespread adoption of AI. For example, due to the potentially debilitating or even fatal consequences of misdiagnosis, the healthcare system has been wary of implementing AI diagnostic algorithms with unknown failure modes. 

One solution, particularly relevant for medicine, is to develop decision support systems with ``humans-in-the-loop''; in this mode, humans treat the output of AI algorithms as ``advice'', while ultimately making the final decision themselves. The AI advice becomes another piece of information to aid in the decision-making process, similar to ordering additional lab tests or consulting a colleague.  
While this mitigates issues related to safety, one of the biggest barriers to adoption remains the black-box nature of current AI systems that limits people's trust in their advice  \cite{feldman2019artificial,ribeiro2016should,xie2020chexplain,miller2019explanation}.

In this paper, we seek to better characterize how humans use advice from an AI algorithm. In particular, we ask whether people use advice differently if it comes from a human versus an AI. To answer this question, we employ the judge-advisor paradigm (JAS) from psychology to compare how people use advice from different sources \cite{van2018advice, prahl2017understanding}. We developed several experiments for a layperson audience and deployed these experiments on a crowdsourcing platform. Our findings suggest that the advice source (human or AI) affects how likely the individual is to use the advice. Individuals use the advice to the extent that they believe the AI or other humans are good at the given task. However, if an individual chooses to use the advice, there is little difference in how they incorporate human advice versus AI advice. We found similar results when we tested this paradigm on experts---dermatologists---who are given the task of deciding whether or not to biopsy a lesion.

\paragraph{Our contributions}
Understanding how users account for AI advice is an important aspect of human-AI interaction and has been under-explored. Here we systematically study how humans use advice in several experiments with both laypeople and an expert group. 
\begin{itemize}
    \item We propose the ``activation-integration'' model for human behaviour in human-in-the-loop systems, where humans first decide whether to use advice and subsequently decide how to update their judgments. 
    \item We quantify how different demographic and task-related factors contribute to a person's response to advice. We identify three key factors related to the task instance -- (1) the person's confidence, (2) the advice confidence, and (3) whether the person's response agrees with the advice -- and one key factor related to the task in general -- the person's prior belief of human vs. AI performance. Other demographic and task-related factors do not have any consistent, significant effect across the variety of settings we test our model under.
    \item We ran extensive experiments to support our hypotheses. We collected data from a variety of tasks, participants living in a variety geographic regions (US, UK, and Asia), and a range of perceptions of advice quality.
    \item We release the data we collected for general use by the research community.\footnote{\href{https://github.com/kailas-v/human-ai-interactions}{https://github.com/kailas-v/human-ai-interactions}}
\end{itemize} 

\paragraph{Related work}
Advice utilization has been studied in the psychology literature where studies often employ JAS for comparing how humans use advice from various sources when completing a given task \cite{van2018advice}. This setup is similar to ours: people are given a task, asked to complete the task, and subsequently receive advice from an outside source. By comparing various types of advice, it is possible to draw conclusions on how different sources of advice are utilized. Previous work has attempted to understand the effects of peer vs. expert advice \cite{madhavan2007effects}, human vs. algorithmic advice \cite{prahl2017understanding, dzindolet2002perceived, madhavan2007effects, onkal2009relative}, task difficulty \cite{gino2007effects}, and advice confidence \cite{sah2013cheap, schultze2015effects} on human utilization of advice. New work has also begun to explore AI advice. In particular, how perception of an AI can affect trust and utilization of its advice \cite{lee2018understanding} and advice utilization of an AI relative to peer or expert humans \cite{tauchert2019following, mesbah2021whose, gaube2021ai}.

The findings so far are inconsistent across settings, suggesting that many aspects of the task and the advice presentation affect advice utilization \cite{jussupow2020we}. Some studies show people utilize automated or AI advice more than advice from peers \cite{dzindolet2002perceived, madhavan2007effects, tauchert2019following, mesbah2021whose}, while others have mixed results where AI advice has less or similar utilization \cite{prahl2017understanding, gaube2021ai}. Most studies use student populations in their experiments, with recruited study participants often US-based. All studies have limited sample sizes, generally with $n \leq 300$. Additionally, studies typically investigate a single data modality like image-based predictions \cite{dzindolet2002perceived, madhavan2007effects, gino2007effects, sah2013cheap, gaube2021ai} or forecasting predictions (including future stock price prediction) \cite{prahl2017understanding, onkal2009relative, tauchert2019following}.

We seek to extend this line of work. We ran a comprehensive set of experiments across multiple geographic regions, data modalities including image-based tasks, text-based tasks, and tabular data-based tasks, and a variety of experiment settings varying participants' beliefs about the quality of the advice given. To run this set of experiments, we recruited over 1100 crowdworkers. This allows us to make more general conclusions about how AI advice is used relative to human advice. In particular, we hypothesize and validate the effectiveness of a two-stage model for human utilization of advice, building on prior work that suggested humans may process advice in two stages \cite{onkal2009relative}. Furthermore, we attribute a human's prior belief of the advice source's suitability for a given task as a key factor for determining how receptive the person will be to advice. We also identify factors that are generally important for advice utilization, supporting findings from previous work \cite{gino2007effects, sah2013cheap, schultze2015effects}.
\section{Study Design}
We ran multiple experiments to investigate how people use AI advice. All of these experiments follow the same basic design.

\begin{figure}[ht]
\centerline{\includegraphics[width=100mm]{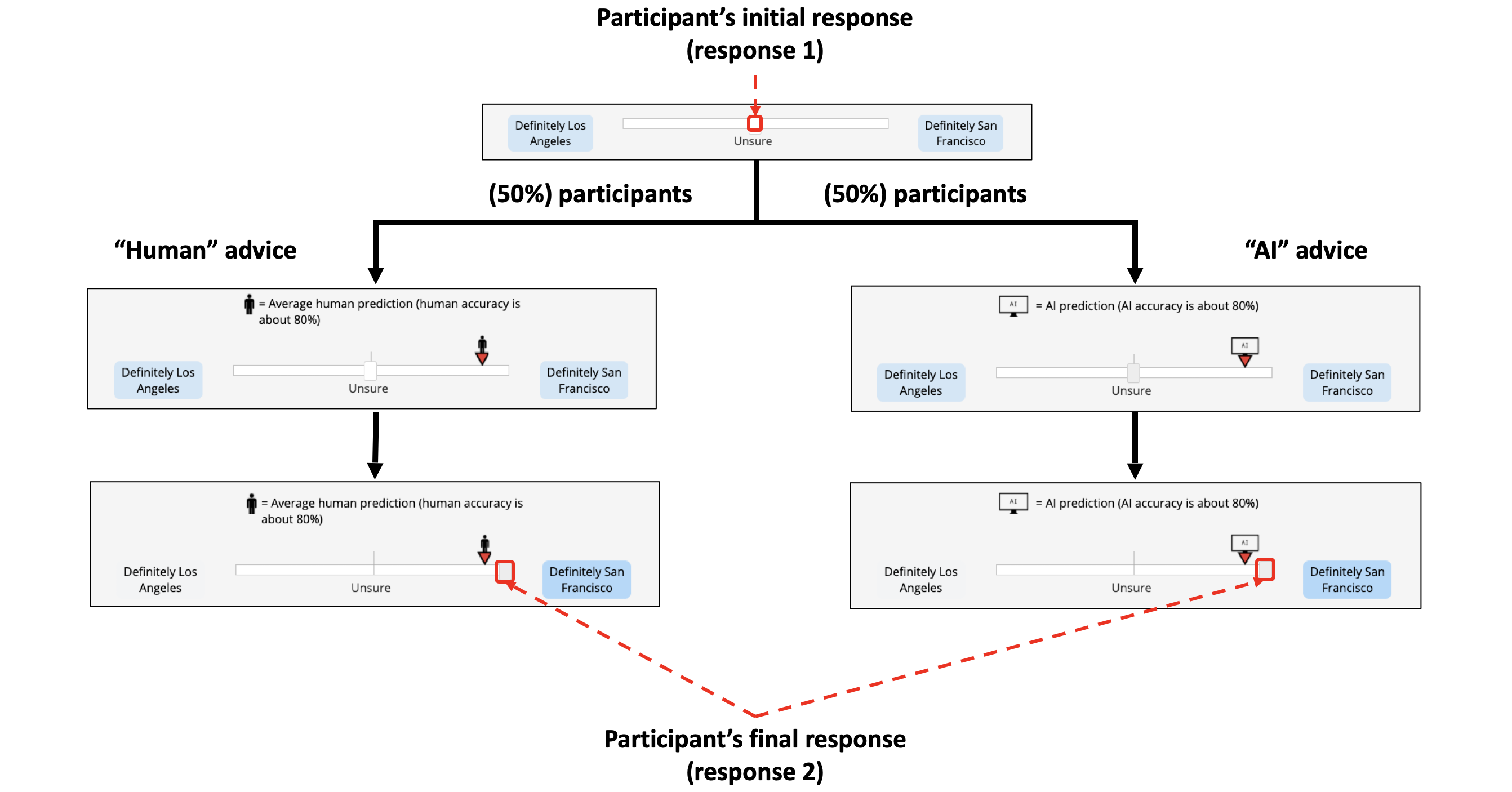}}
\caption{
Visualization of our study design. 
}
\label{fig:study_design}
\end{figure}

\subsection{Overview}
To investigate whether participants use advice from an AI algorithm differently than from humans, we developed a set of experiments on a variety of tasks (Section~\ref{sec:tasks_overview}). 
The experiments are setup as the 2-stage procedure shown in Figure~\ref{fig:study_design}. Participants were recruited from a crowdsourcing platform and randomly assigned into two groups. The first group received advice from a human while the second group received advice from an AI algorithm. Participants were adults from the US. We also recruited participants from the UK and Asia for a subset of the experiments (Section~\ref{sec:regression_global}). 

To isolate the advice source's effect on people's judgments, the AI-labeled advice and human-labeled advice were identical. The advice was generated using the average response from a previous set of $\sim$50 human labelers who completed the tasks without advice. The resulting advice has an accuracy of $80-90\%$, depending on the task. 
In each condition, participants were informed that the advice source was 80\% accurate, and the actual advice participants received was identical in both conditions. Between the conditions, we only varied whether the advice was given by a human versus AI and how the advice was presented by using a small icon of a person or a computer (see Figure~\ref{fig:study_design}). When showing the advice to participants, we randomly perturbed the advice by 3 fixed amounts: (1) $20\%$ of the scale width away from the correct response, (2) $20\%$ of the scale width towards the correct response, or (3) no change. Options (1) and (2) each had a $25\%$ probability of being selected, while option (3) had $50\%$ probability of being selected. We added these perturbations as a means for both increasing the variety of advice given and decreasing the accuracy of the advice to $<=$80\%. The expected advice accuracy is given in Table~\ref{tbl:datasets_summary}. The randomness in these perturbations is specific to each participant. Note that participants did not receive feedback on their or the advice's performance on individual questions during the experiment.

After reading through instructions, each participant completed a series of tasks in randomized order. Within each experiment, the tasks were all of the same type (e.g., image classification on city images), but the difficulty of each task instance could vary significantly (see Table~\ref{tbl:datasets_summary}). All tasks are binary classification tasks. Users submitted answers in the form of a confidence on a continuous sliding scale with endpoints labeled ``Definitely [Label A]'' and ``Definitely [Label B]''.

Participants were incentivized to perform well on the task by receiving a bonus based on their averaged judgments across all task instances (both before and after advice). Participants were informed of how the bonus was calculated in the instructions. After completing all tasks in an experiment, participants were asked a series of survey questions to gauge their trust in and familiarity with AI. More details on the instructions and the survey questions are provided in Appendix~\ref{sec:survey}.

\subsection{Survey Question About Prior} \label{sec:prior_survey}
The prior belief value reported in Table~\ref{tbl:datasets_summary} was obtained using a survey question asked to every participant after they had finished all task instances. We also ran additional experiments to verify that the location of this survey question had no affect on our analysis and conclusions (see Appendix~\ref{app:prior_survey}). The question asked is shown below. The wording was slightly modified for those who received AI advice or human advice to account for the advice they received. The ordering of AI and human in the question was randomized for each participant.

\paragraph{Human:} ``Do you think an artificial intelligence (AI) algorithm or the average person (without help) can do better on this task?''
\paragraph{AI:} ``Do you think the AI or the average person (without help) can do better on this task?''

\subsection{Tasks}\label{sec:tasks_overview}

\begin{figure}
\begin{tabular}{c}
  \includegraphics[width=55mm]{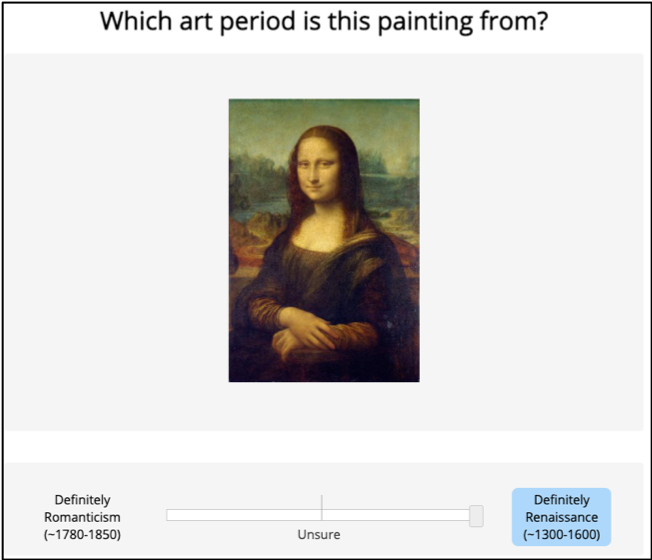} \\ (a) Art \\[1pt] \includegraphics[width=55mm]{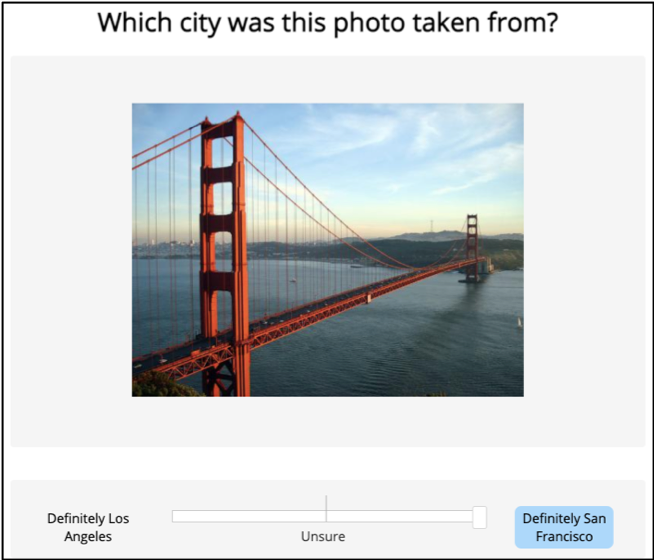} \\
 (b) Cities \\[1pt]
 \includegraphics[width=55mm]{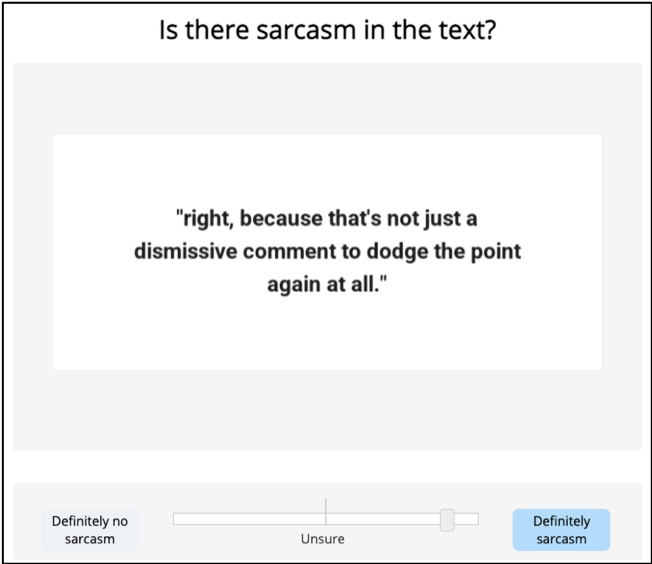} \\ (c) Sarcasm  \\[1pt] \includegraphics[width=55mm]{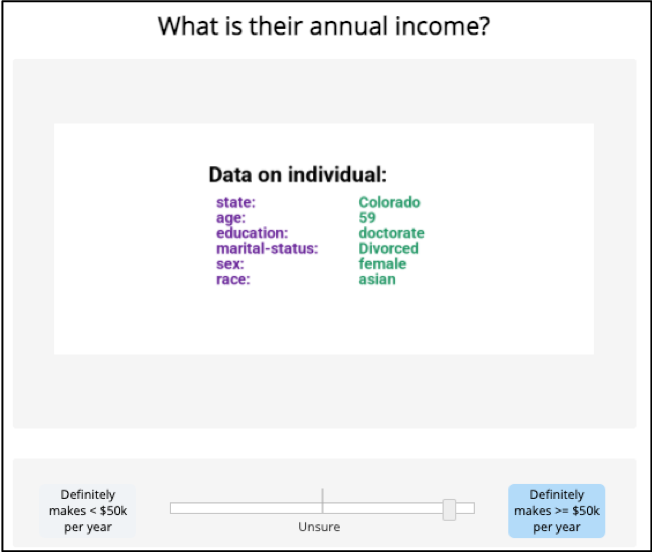} \\
 (d) Census \\[1pt]
\end{tabular}
\caption{Example tasks for each of the 4 datasets we use.}
\label{fig:tasks}
\end{figure}

\begin{table*}[h]
\caption{Summary of our 4 laypeople datasets. Baseline accuracy describes task instance-level performance prior to advice (average +/- standard deviation). Prior belief indicates whether participants favored AI or humans; see Section~\ref{sec:prior_survey} for details.}
{\resizebox{2\columnwidth}{!}{\begin{tabular}{@{}lllcccl}
\toprule
Dataset & Data Type & Description & \# of Tasks & Baseline accuracy & Advice accuracy & Prior belief \\
\midrule
Art Dataset & Image & identify the art period & 32 & 65.7\% +/- 17.0\% & 76.6\% & 66.7\% (AI) \\
Cities Dataset & Image & identify the city & 32 & 72.1\% +/- 17.1\% & 80.5\% & 62.0\% (AI) \\
Sarcasm Dataset & Text & identify sarcasm & 32 & 72.9\% +/- 20.3\% & 78.9\% & 80.0\% (human) \\
Census Dataset & Tabular & identify income level & 32 & 70.4\% +/- 28.8\% & 73.4\% & 66.0\% (AI) \\
\bottomrule
\end{tabular}}
}\label{tbl:datasets_summary}
\end{table*}

We provide an overview of the datasets we use for our experiments in Table~\ref{tbl:datasets_summary}. The socioeconomic score and education level are provided by the crowdsource platform we use for running our experiment (Prolific \cite{prolific}). For each experiment, every participant completes all tasks from a given dataset twice -- before and after receiving advice.
More details about each dataset are given below. In Figure~\ref{fig:tasks}, we show sample tasks from each dataset. These were the examples shown to participants in the instructions and were selected to be easier than the tasks we use in our experiments. The tasks were designed to cover diverse data modalities (visual, text and tabular) as well as sufficiently challenging so that participants can benefit from advice.

\paragraph{Art dataset (Image)}
This dataset contains images of paintings from 4 art periods: Renaissance, Baroque, Romanticism, and Modern Art. The dataset contains 8 paintings from each time period. Participants were asked to determine the art period a painting is from given a binary choice. The incorrect label was selected to be from adjacent time periods to increase task difficulty (e.g., if Romanticism is the correct label, then the alternate choice would be Baroque or Modern Art; this choice was fixed for all participants). An example task is shown in Figure~\ref{fig:tasks}a.

\paragraph{Cities dataset (Image)}
This dataset contains images from Google from 4 major US cities: San Francisco, Los Angeles, Chicago, and New York City. The dataset contains 8 images from each city. The task is to identify which city an image is from given a binary choice. For each image, an alternate city was randomly selected as the negative label presented to participants. The images were selected to be moderately difficult to identify -- images of major landmarks were excluded, but differentiating aspects of the cities like architecture were included. An example task is shown in Figure~\ref{fig:tasks}b. 

\paragraph{Sarcasm dataset (Text)}
This dataset is a subset of the Reddit sarcasm dataset \cite{khodak2017large}, which includes text snippets from the discussion forum website, Reddit. Specifically, the subset was selected from posts in the ``AskReddit'' forum and were hand-filtered to exclude posts that either contained potentially offensive content or were too long. Survey participants were asked to identify sarcasm in the text snippet. The dataset was balanced to contain 16 sarcastic posts and 16 non-sarcastic posts. An example task is shown in Figure~\ref{fig:tasks}c.

\paragraph{Census dataset (Tabular)}
This dataset comes from a subset of US census data \cite{census2019}. The task is to identify an individual's income level (does the individual make $>=\$50,000$ annual income or not?) given their demographic information: state of residence, age, education level, marital status, gender, and race. These features were selected based on previously published algorithms for predicting income level \cite{Dua:2019,chakrabarty2018statistical}. The dataset was balanced to contain 16 individuals who make $>=\$50,000$ and 16 who make $<\$50,000$ annual income. The dataset was further balanced across race and gender and each binary income category contains 4 Asian, 4 Black, and 8 White individuals. An example task is shown in Figure~\ref{fig:tasks}d. 

The census dataset may raise concerns of reinforcing gender or race-related biases in participants. However, as income levels were balanced across race and gender, biased participants would actually receive negative feedback as they would necessarily receive a low bonus payment due to their bias.
\section{Results}\label{sec:results}
In Section~\ref{sec:results-effect_of_advice}, we establish that both AI and human advice are helpful across our tasks. In Section~\ref{sec:results-ai_vs_human}, we identify the key difference between human and AI advice and propose our \textit{activation-integration model} to model the observed human behavior. In Section~\ref{sec:prior_effect}, we demonstrate the reason for a difference between human and AI advice is largely due to the prior beliefs of an individual. Finally, we validate our activation-integration model and prior belief results across geographic locations and study conditions in Section~\ref{sec:validation_studies} and with expert participants using a dermatology task in Section~\ref{sec:biopsy_task}.

The results in Sections~\ref{sec:results-effect_of_advice}, \ref{sec:results-ai_vs_human}, and \ref{sec:prior_effect} are based on the data summarized in Table~\ref{tbl:datasets_participant_summary}. The data used in Sections~\ref{sec:validation_studies} and \ref{sec:biopsy_task} are described in those sections.

\subsection{Effect of advice}\label{sec:results-effect_of_advice}

\begin{table*}[h]
\caption{Participant demographics for our initial experiments. Socioeconomic score ($[1-10]$): 5 corresponds to a middle class. Education level ($[1-8]$): 5 corresponds to an associate's degree. See Appendix~\ref{sec:demographic_details} for details.
}
{\resizebox{2\columnwidth}{!}{\begin{tabular}{@{}lM{7em}M{7em}M{7em}M{7em}M{7em}r@{}}
\toprule
Dataset & \# of Participants & Gender (Percent Male / Female) & Age            & Socioeconomic Score & Education Level \\ 
\midrule 
Art     & 147                & 47.6\% / 52.4\%               & 33.0 +/- 12.3  & 5.2 +/- 1.5         & 5.4 +/- 1.3 \\
Cities  & 94                 & 50.0\% / 50.0\%               & 26.1 +/- 7.8   & 5.3 +/- 1.5         & 5.4 +/- 1.2 \\ 
Sarcasm & 97                 & 43.3\% / 56.7\%               & 31.2 +/- 11.7  & 5.3 +/- 1.6         & 5.2 +/- 1.4 \\
Census  & 98                 & 46.9\% / 53.1\%               & 30.5 +/- 10.9  & 5.2 +/- 1.6         & 5.3 +/- 1.2 \\ 
\bottomrule
\end{tabular}
}}\label{tbl:datasets_participant_summary}
\end{table*}

\begin{table}[h]
\caption{
Treatment effect (Sections~\ref{sec:results-effect_of_advice}, \ref{sec:results-ai_vs_human}, and \ref{sec:prior_effect}). 
Accuracy $\Delta$ is the accuracy change after receiving advice. 
``Activation Rate'' is the percentage of participants who change their response after receiving advice. 
Activation $\Delta$ is the difference between AI and human activation.
}
{\resizebox{\columnwidth}{!}{\begin{tabular}{@{}lM{4em}M{4em}M{4em}M{4em}@{}}
\toprule
Dataset & Art & Cities & Sarcasm & Census \\
\midrule
Accuracy \\
\midrule
Before Advice & 65.7\% & 72.5\% & 73.1\% & 70.5\% \\
$\Delta$ Human & +7.1\% & +3.7\% & +3.3\% & +2.4\% \\
$\Delta$ AI & +11.8\% & +6.2\% & +3.3\% & +3.5\% \\
\midrule
Activation Rate \\
\midrule
Human & 46.0\% & 48.2\% & 39.8\% & 41.8\% \\
AI & 52.3\% & 54.5\% & 34.0\% & 47.5\% \\
$\Delta$ & +6.4\% & +6.3\% & $-$5.8\% & +5.7\% \\
\bottomrule
\end{tabular}
}}\label{tbl:datasets_results_summary}
\end{table}

\begin{figure*}
\begin{tabular}{cc}
  \includegraphics[width=65mm]{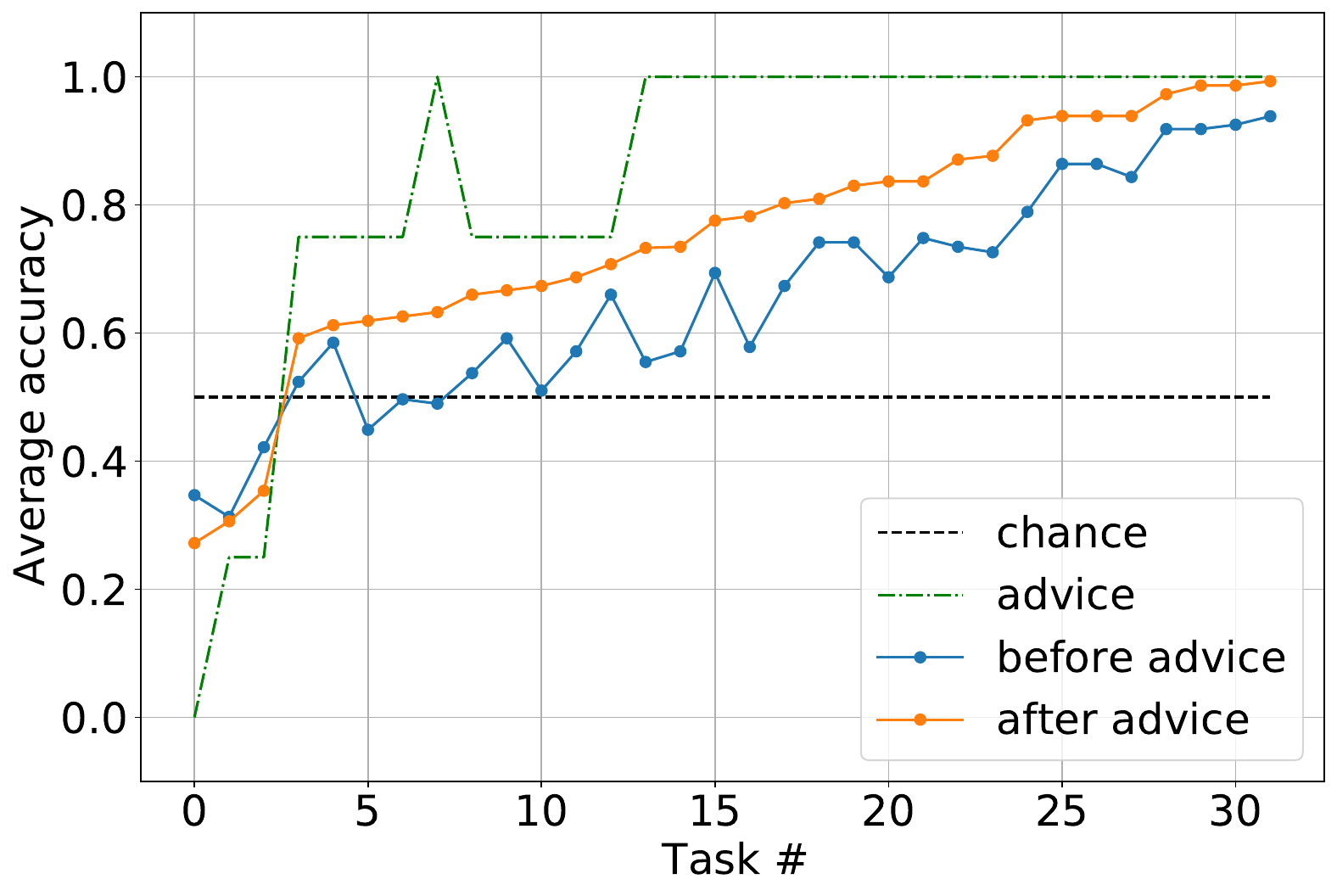} &   \includegraphics[width=65mm]{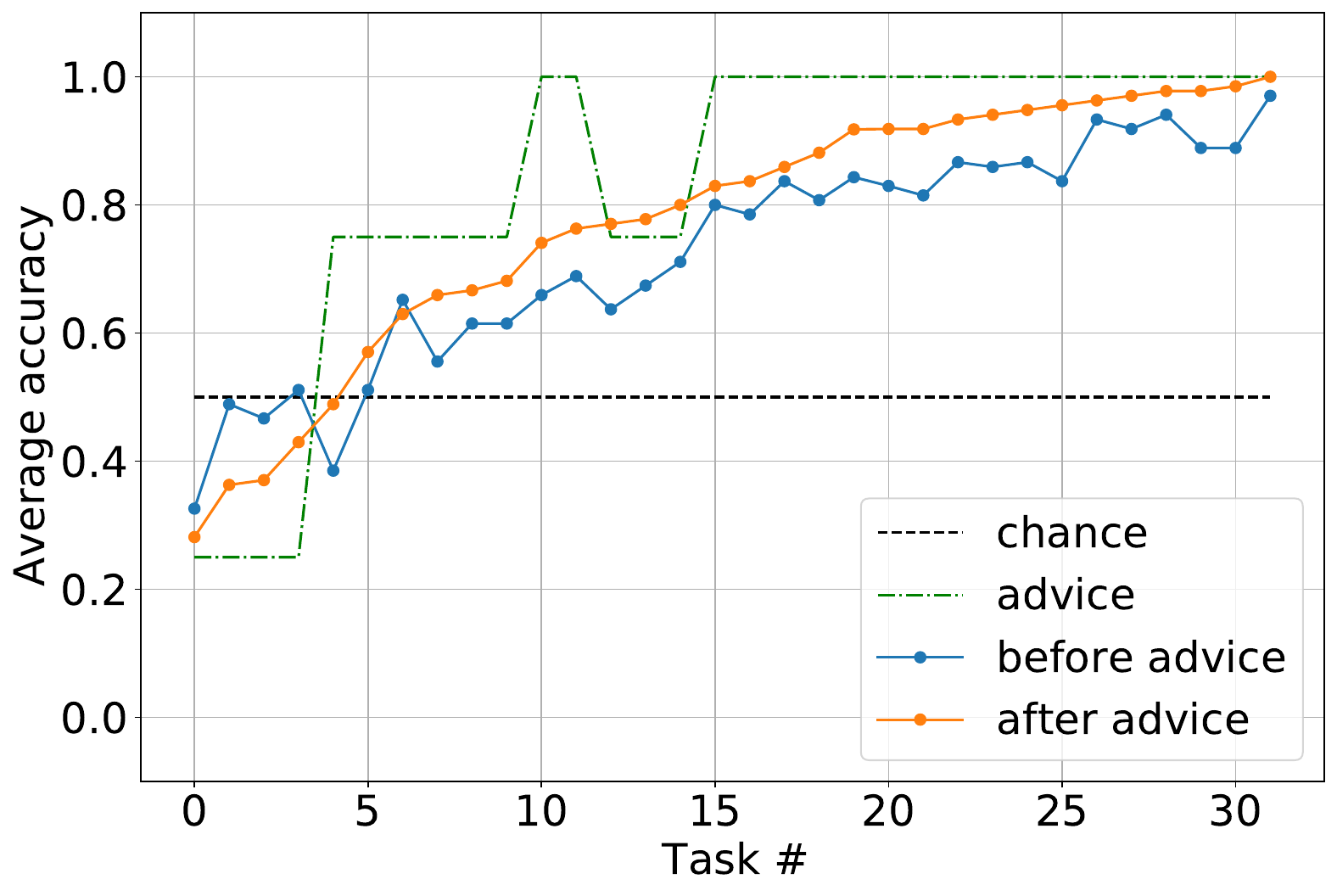} \\
(a) Art & (b) Cities \\[6pt]
 \includegraphics[width=65mm]{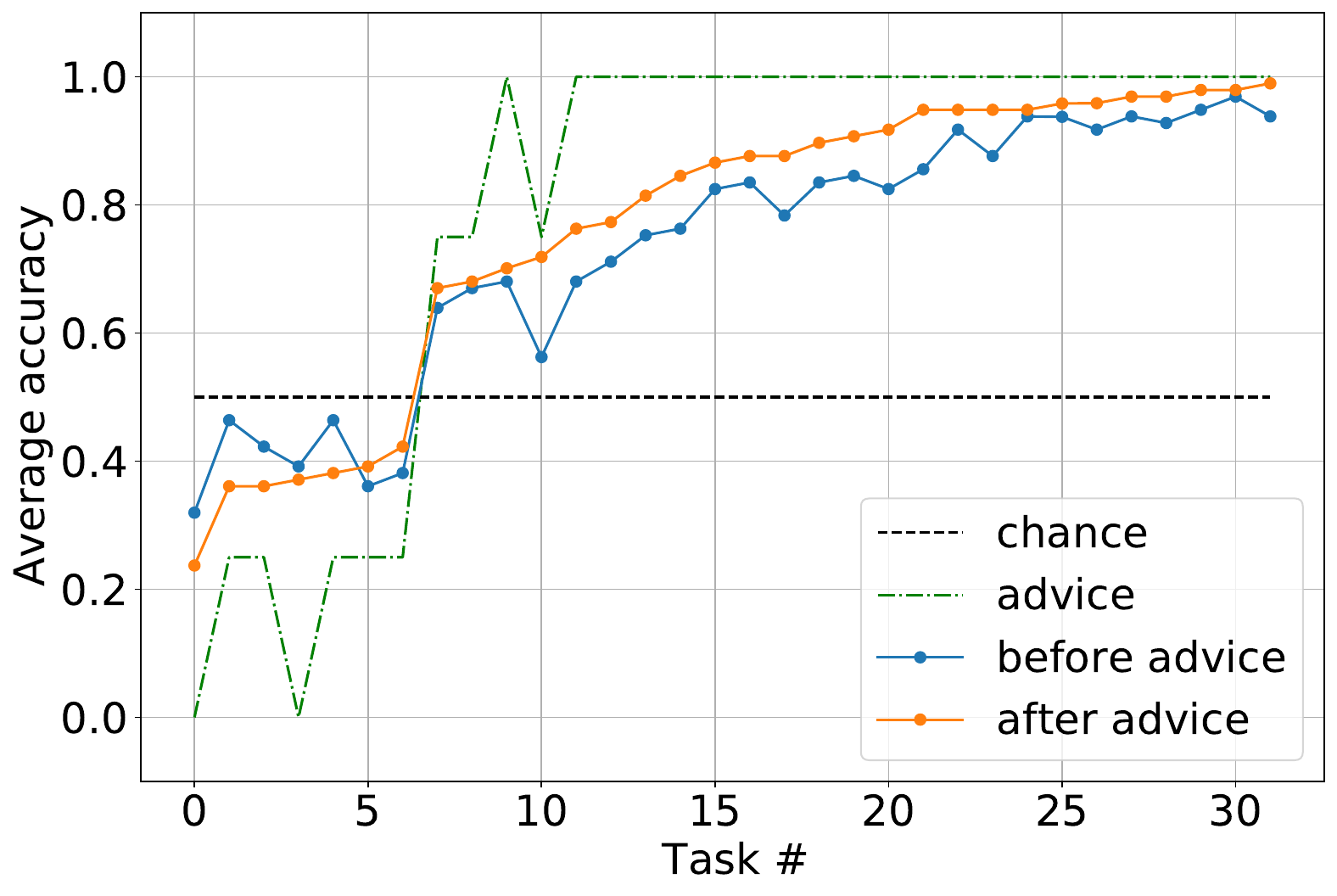} &   \includegraphics[width=65mm]{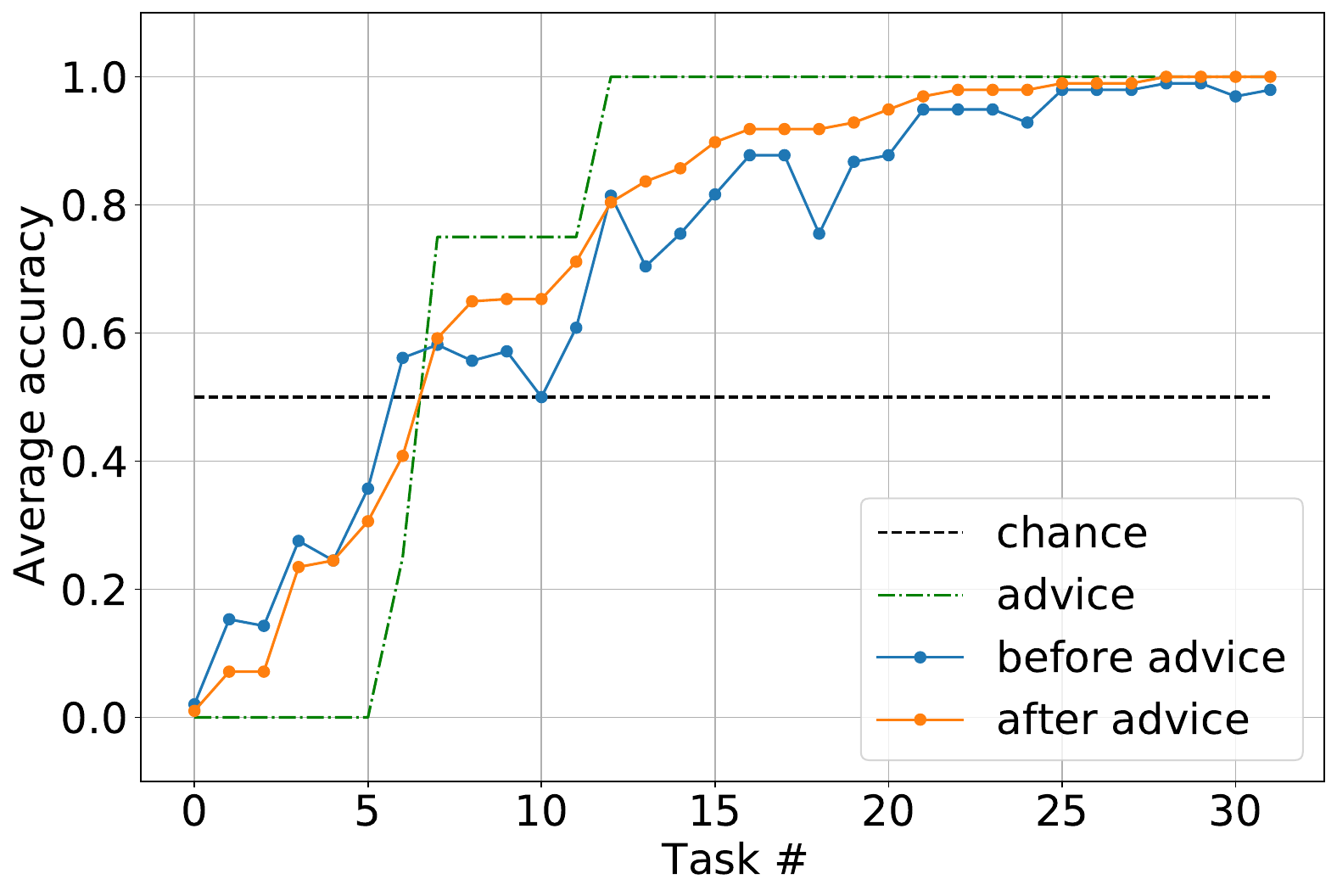} \\
(c) Sarcasm & (d) Census \\[6pt]
\end{tabular}
\caption{
Average accuracy before and after receiving advice. \textit{chance} is the accuracy expected from random guessing; \textit{advice} is the average accuracy of the advice; \textit{before advice} and \textit{after advice} are the average accuracies across both arms of the experiment.
}
\label{fig:advice_effect}
\end{figure*}

First, we consider the effect of receiving any advice on performance. This effect is summarized in the \nth{1} through \nth{3}  rows of Table~\ref{tbl:datasets_results_summary}. The \nth{1} row reports the average accuracy across tasks before receiving advice, and the \nth{2} and \nth{3} rows report the average change in accuracy after receiving advice. In all four datasets, participants' responses are more accurate after having received advice in both human and AI conditions. We visualize the average accuracy difference across individual task instances in Figure~\ref{fig:advice_effect}. 

\subsection{Difference between AI and human advice}\label{sec:results-ai_vs_human}
\begin{figure*}
\begin{tabular}{cc}
  \includegraphics[width=65mm]{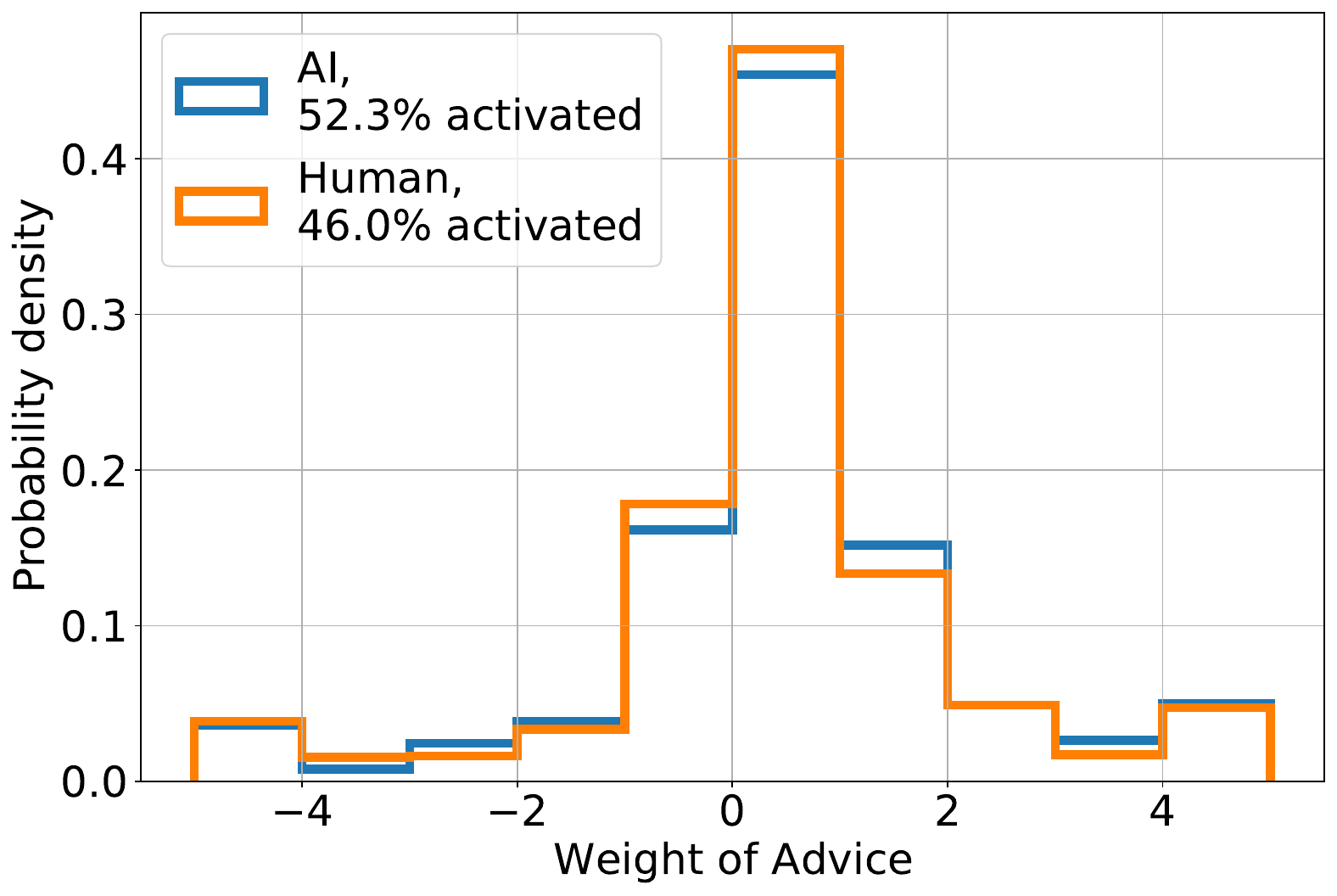} &   \includegraphics[width=65mm]{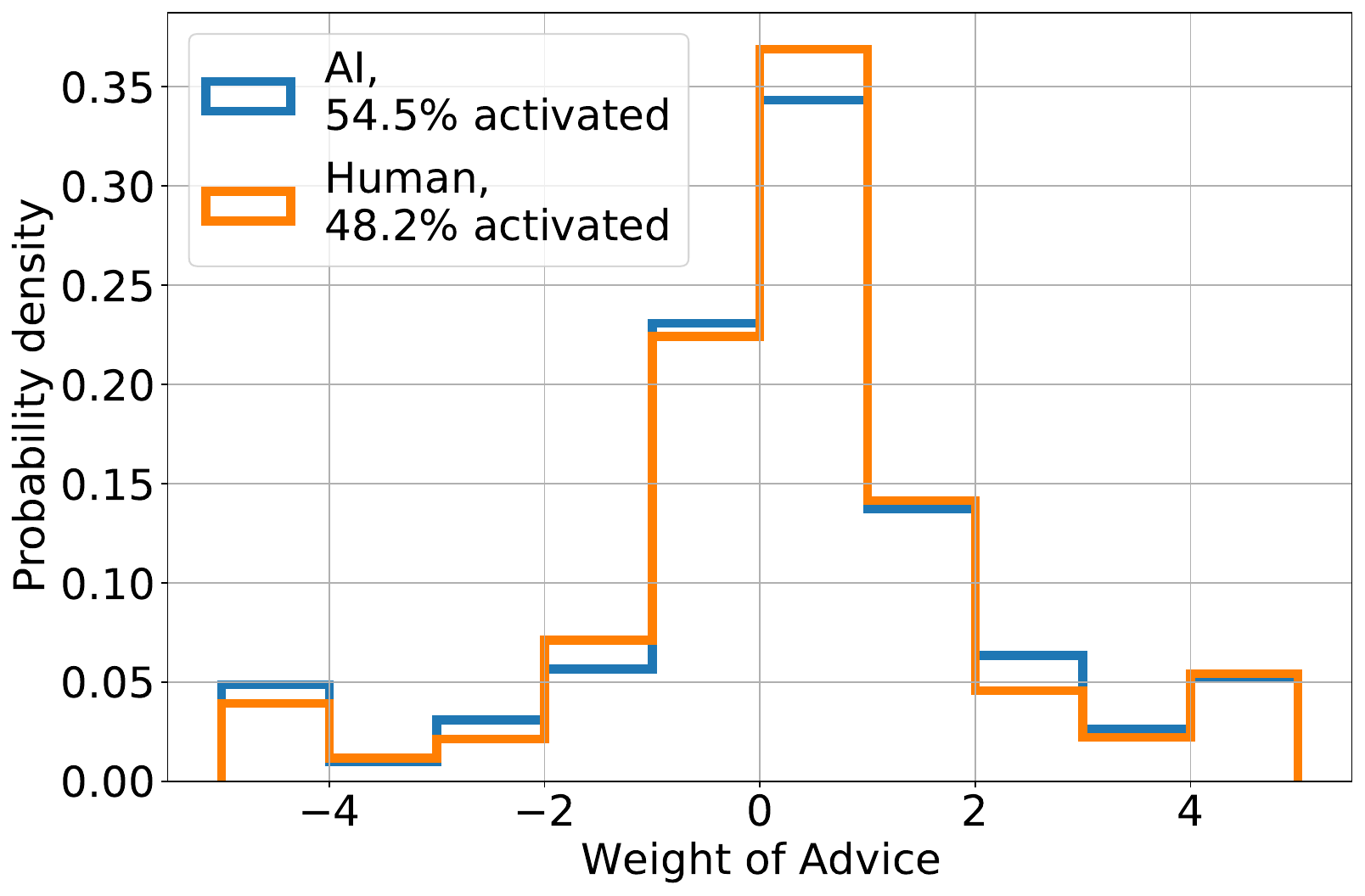} \\
(a) Art & (b) Cities \\[6pt]
 \includegraphics[width=65mm]{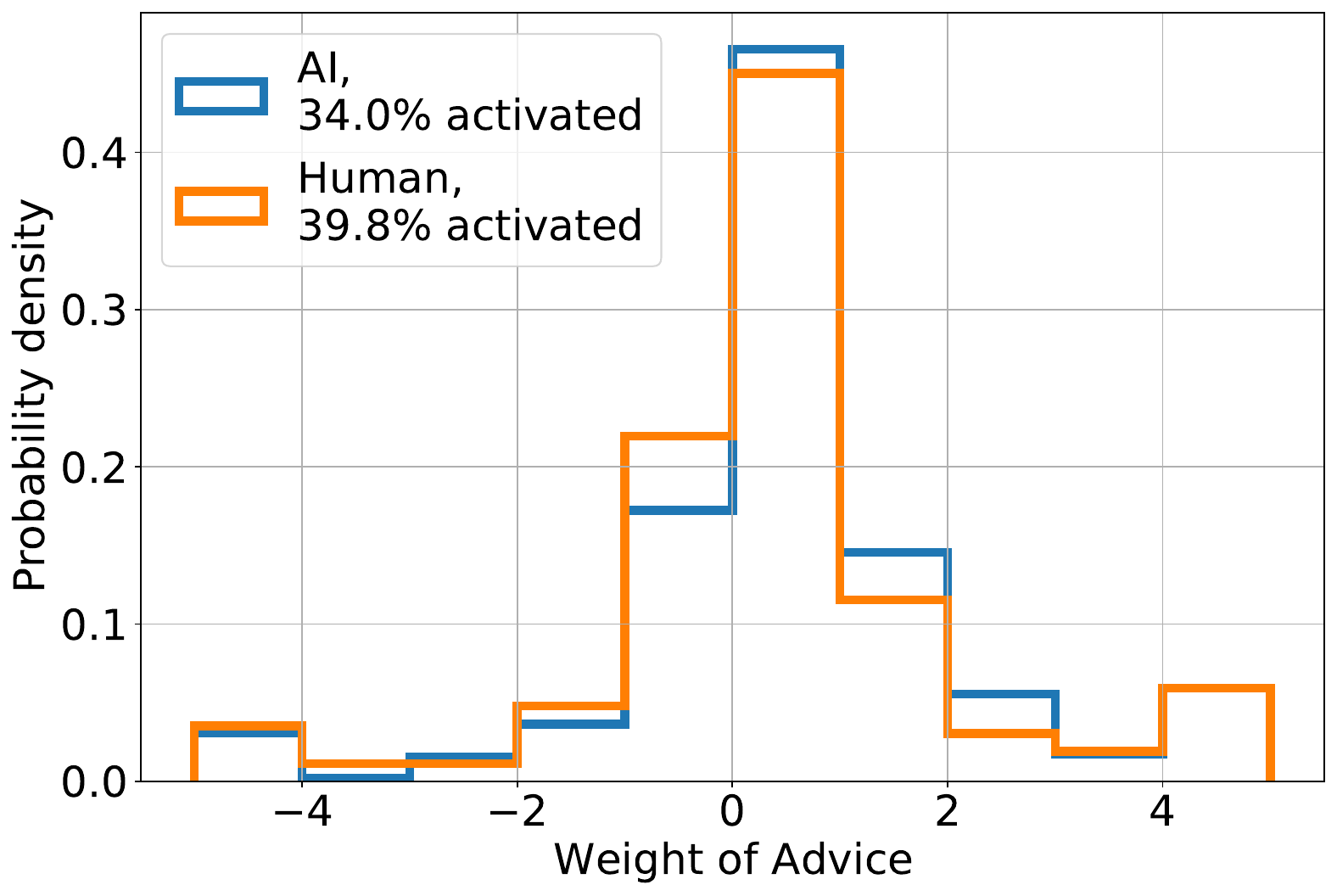} &   \includegraphics[width=65mm]{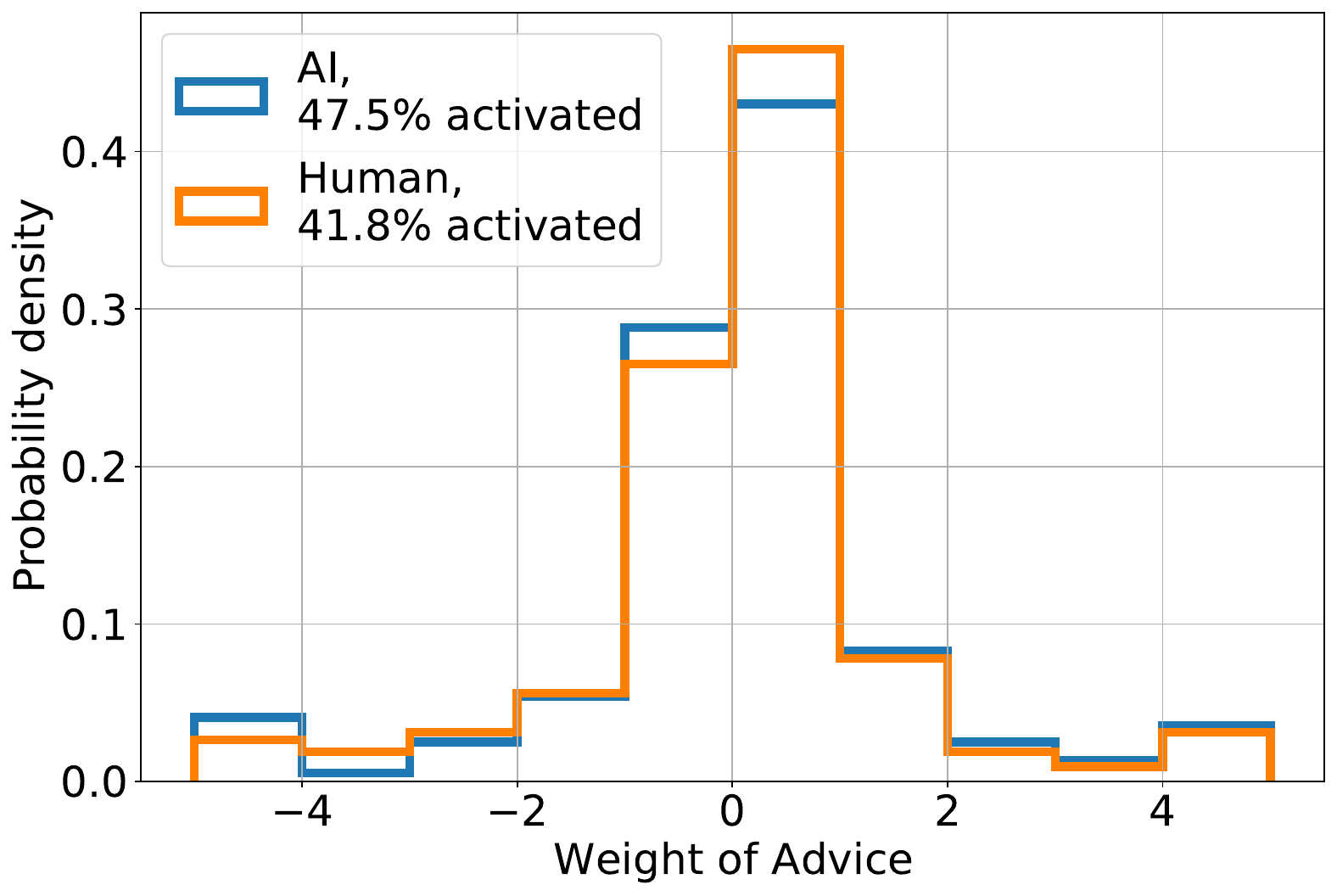} \\
(c) Sarcasm & (d) Census \\[6pt]
\end{tabular}
\caption{
PDF of the weight of advice (WoA), conditioned on being activated. WoA is clipped to have a maximum magnitude of 5.
The $\%$ activated refers to the corresponding value in Table~\ref{tbl:datasets_results_summary}.
}
\label{fig:woa}
\end{figure*}

Now we ask how this effect differs between human versus AI advice. To help answer this question, we will use the weight of advice (WoA), a common metric in the psychology of advice utilization \cite{harvey1997taking, yaniv1997precision}. The WoA is defined as 
\begin{equation} \label{eq:woa}
    \text{WoA} = \frac{\text{response}_2 - \text{response}_1}{\text{advice} - \text{response}_1},
\end{equation}
where $\text{response}_1$ and $\text{response}_2$ are the participant's first and second answers respectively, and $\text{advice}$ is the advice presented to the participant. WoA is typically clipped to have bounded magnitude (e.g., in the case that the first response happens to be close to the advice given). In the literature, WoA is typically clipped to have a maximum magnitude of $1$ (i.e., perfectly following the advice). However, we opt to clip the WoA to have maximum magnitude of $5$, allowing us to capture the possibility that people become more confident than the advice after integrating with their own knowledge. A positive value implies that a participant changed their response in accordance with the advice, while a negative value implies the participant changed their response away from the advice. A response of 0 occurs when the participant did not change their response.  We define a user to be ``activated'' for a given task if they change their response by at least a threshold amount after receiving advice. This threshold was chosen to be equal to the width of the slider knob ($3.5\%$ of the total slider bar width).

We plot the probability density function of the WoA for each dataset, comparing across conditions in Figure~\ref{fig:woa}. From these plots, we make the following observations:

\begin{enumerate}
    \item[(1)] There is a statistically significant difference in how often participants modify their initial response when given AI versus human advice. The magnitude and direction of this difference varies between datasets. For Art, Cities, and Census, the AI condition evokes a larger response; for Sarcasm, the human condition evokes a larger response. We describe this effect as an ``activation difference,'' where the condition that people are more likely to use advice from is the more activated condition. The magnitude and direction of this difference is shown in the ``$\Delta$'' row of the Activation Rate sub-table in Table~\ref{tbl:datasets_results_summary}.
    
    Also note that the sign of the activation difference is reflected in the change in accuracy across the human and AI subsets. For the datasets where AI had a higher activation rate, the participants receiving AI advice had a greater average change in accuracy. On the sarcasm dataset, the change in accuracy is identical between those receiving human versus AI device. While this may seem contradictory (there is a non-zero activation difference), Figure~\ref{fig:advice_effect} provides an explanation. Note that in the Sarcasm dataset, there are more tasks where the advice is incorrect than in the other three datasets. So, the accuracy improvement we would expect to see is mitigated by bad advice decreasing accuracy. This explanation can also be applied to the Census dataset which has a smaller accuracy difference between AI and human than the Art or Cities dataset.

    \item[(2)] Conditioned on having changed their label, people behave similarly. As shown in Figure~\ref{fig:woa}, this subset of the population corresponds to between $34\%$ and $54.5\%$ of the responses depending on the experiment. We ran a 2 sample Kolmogorov–Smirnov test for each set of tasks with a p-value threshold of 0.01, including all responses where the label was changed after receiving advice and splitting on the treatment arm to get 2 samples. In all 4 datasets, there was no statistically significant difference between the samples, suggesting the main effect of the treatment is to modulate the activation rate. 
\end{enumerate}

\paragraph{Activation-integration model} Based on these experiments, we propose a two stage model, which we term the activation-integration model, for how participants utilize the advice they receive. In the first (activation) stage, a participant decides whether or not to use the advice. In the second (integration) stage, they decide to what extent they will use the advice. There is evidence in the literature for such a model. Prior work has found evidence that humans process advice in two stages \cite{onkal2009relative}. There is also evidence that people are better at assessing advice than using it \cite{harvey2000using}. Our activation-integration model captures this effect nicely. 

Using this model, we can understand that receiving AI versus human advice affects the activation stage, but not the integration stage. In particular, the rate at which people utilize advice is modulated by the advice source, but the extent to which people utilize advice is not significantly different if they decide to use the advice.

\subsection{Effect of participants' prior beliefs} \label{sec:prior_effect}

\begin{table}[h]
\caption{
Confidence levels of (1) a logistic regression mixed effects model predicting activation and (2) a linear regression mixed effects model predicting integration response. Coefficients printed bold if non-zero (significant with $\alpha=0.95$). $\beta_{\text{consistent}}$ refers to whether the initial response and the advice favor the same label. The $\beta_{\text{response}}$ and $\beta_{\text{advice}}$ refer to the magnitude of the response and advice respectively.
}
{\resizebox{\columnwidth}{!}{\begin{tabular}{@{}M{10em}M{7em}M{7em}@{}}
\toprule
Covariates & Activation Coefficients & Integration Coefficients \\
\midrule
Given AI Advice? ($\beta_{\text{ai}}$) & 0.162 +/- 0.134 & -0.023 +/- 0.016 \\
Prior Belief ($\beta_{\text{prior}}$) & \textbf{0.244 +/- 0.059} & 0.001 +/- 0.008 \\
Response 1 Confidence ($\beta_{\text{response}}$) & \textbf{-0.584 +/- 0.025} & \textbf{-0.212 +/- 0.004} \\
Advice Confidence ($\beta_{\text{advice}}$) & \textbf{0.218 +/- 0.024} & \textbf{0.025 +/- 0.004} \\
Response 1 and Advice Consistent? ($\beta_{\text{consistent}}$) & \textbf{-1.555 +/- 0.051} & \textbf{0.661 +/- 0.008} \\
Age ($\beta_{\text{age}}$) & -0.089 +/- 0.072 & \textbf{-0.026 +/- 0.008} \\
Gender ($\beta_{\text{gender}}$) & -0.046 +/- 0.137 & 0.010 +/- 0.016 \\
Education ($\beta_{\text{education}}$) & 0.043 +/- 0.076 & \textbf{0.021 +/- 0.009} \\
Socioeconomic Status ($\beta_{\text{socioeconomic}}$) & \textbf{0.139 +/- 0.070} & 0.007 +/- 0.008 \\
Has Programming Experience? ($\beta_{\text{programming}}$) & 0.233 +/- 0.147 & 0.014 +/- 0.017 \\
\bottomrule

\end{tabular}
}}\label{tbl:regression_coefficients}
\end{table}

Now we seek to explain what accounts for the treatment effect we observed. The final column of Table~\ref{tbl:datasets_summary} offers an explanation: it shows participants' answers to a survey question about whether an AI algorithm or the average human would perform better on the given task. Responses were recorded on a sliding scale with human at one end and AI at the other end. We computed the prior by averaging all responses and rescaling to a range of $-100\%$ to $+100\%$. We then set the sign such that a positive prior corresponds to the participant's belief that the advice source they received would perform better than the alternate source. We visualize the full distribution in Appendix~\ref{sec:prior}.

We use 2 mixed effects linear models to test our activation-integration model and show the effect of the prior. The list of covariates used as fixed effects is included in Table~\ref{tbl:regression_coefficients}; we use task ID and subject ID for the random effects as random intercepts. A mixed effects model has the form 
\begin{equation} \label{eq:mixed_effects_model}
    y = \beta X + \mu Z + \epsilon,
\end{equation}
where $y$ is the outcome variable, $X$ is a matrix of the covariates of interest (the fixed effects), $\beta$ is the coefficients for the covariates, $Z$ is the random effects design matrix, $\mu$ is the coefficients for the random effects, and $\epsilon$ is additive noise. $\mu$ is a Gaussian random variable that, in our case, models variations across participants and stimuli. As per standard practice, we include random intercepts for both random effects \cite{barr2013random}.
Continuous covariates were normalized to have $0$ mean and unit variance. We jointly fit the model across all four datasets using the lme4 R package \cite{bates2014fitting}.

In the activation stage, we use activation on a task as our outcome variable. As this is a binary variable, we use a logistic regression mixed effects model. We list the fitted coefficients in the ``Activation Coefficients'' column of Table~\ref{tbl:regression_coefficients}. There are a few key takeaways. The first is that the  belief about AI versus human performance on a task ($\beta_{\text{prior}}$) has a significant effect on whether an individual is activated. The sign of $\beta_{\text{prior}}$ indicates that participants who received advice from the source they believe is better for the task are more likely to be activated. Other important coefficients include confidence of response  ($\beta_{\text{response}}$) -- its negative value suggests that unsure participants are more inclined to be activated and advice confidence ($\beta_{\text{advice}}$) -- its positive value suggests that participants are more likely to activate when presented more confident advice. While one might think agreement between the response and advice ($\beta_{\text{consistent}}$) should have a positive coefficient, it is negative which suggests that when a participant's initial response agrees with the advice, they are less likely to activate. For example, because the response scale has a fixed width, participants who are already very sure about their response ($96.5\%$ or greater) cannot move the knob any further when the advice agrees with them.  Thus, these individuals cannot be "activated".

To model the integration stage, we use change in response after advice (sign-aligned so that $>0$ corresponds to greater certainty in the original label prediction) as our outcome variable. This outcome is a continuous variable and so we use a linear mixed effects model. We list the learned coefficients in the ``Integration Coefficients'' column of Table~\ref{tbl:regression_coefficients}. There are three key takeaways. $\beta_{\text{response}}$ has a large negative coefficient suggesting, similar to the activation model, that participants sure about their response do not change it as much. $\beta_{\text{consistent}}$ has a large positive coefficient here -- agreement between the participant's initial response and the advice encourages larger changes. Lastly, note that $\beta_{\text{prior}}$ is not associated with response after advice. This suggests that the primary effect of the prior is determining whether a person activates.

\subsection{Additional validation studies} \label{sec:validation_studies}

\subsubsection{Varying geographic location} \label{sec:regression_global}
As perception of AI may be tied to geographic location \cite{hagerty2019global}, it is important to question whether our results generalize to across geographic regions. To answer this question, we recruited approximately 200 participants from the UK and 200 participants from Asia and ran experiments with our art and sarcasm datasets. Note that the studies were run identically to those given to US-based participants, with the exception that we now recruit participants exclusively from the corresponding geographic region. More details including participant demographics are included in the Appendix~\ref{sec:UKAsia}. 

The main finding is that our activation-integration model is largely validated in the UK and in Asia. In the activation model, the prior is associated with activation in the UK. $\beta_{\text{prior,UK}} = 0.217$ is similar to the $\beta_{\text{prior,US}} = 0.244$, suggesting the two geographical groups are similar. While we do not have sufficient power to say $\beta_{\text{prior,Asia}}=0.101$ is associated with activation, its positive value is in line with the UK and US prior coefficients. $\beta_{\text{response}}$ (US:$-0.584$, UK:$-0.491$, Asia:$-0.355$), $\beta_{\text{advice}}$ (US:$0.218$, UK:$0.146$, Asia:$0.143$), and $\beta_{\text{consistent}}$ (US:$-1.555$, UK:$-1.157$, Asia:$-1.492$) are associated with activation and are similar in sign and magnitude across all three regions. Gender, education, age, and programming experience are not associated with activation probability across all three regions. 

In the integration model, $\beta_{\text{response}}$ (US:$-0.212$, UK:$-0.199$, Asia:$-0.226$) and $\beta_{\text{consistent}}$ (US:$0.661$, UK:$0.589$, Asia:$0.674$) are associated with integration and have similar sign and magnitude across all three regions. Neither treatment nor prior are associated with integration across all three regions, validating our earlier conclusion that the prior has a strong effect in determining activation, but a weak effect in utilization of advice after activation.

\subsubsection{Varying perceived accuracy} \label{sec:perceived_acc}
We also quantified the effect of the advice's perceived accuracy on its utilization. In previous experiments, we informed participants that the advice has an accuracy of $80\%$, for both AI and human advice arms. Here, we varied this value to $65\%$ and $95\%$. Note the actual advice given to participants is the same across the three groups -- we only varied the advice description. 
In this experiment, we focused only on the art dataset. We re-used the previously obtained data for the $80\%$ group and collected new data for the $65\%$ and $95\%$ groups. All data were collected with US participants. More details including participant demographics are included in Appendix~\ref{sec:perceived_acc_val}. 

We observe the previous findings relating to our activation-integration model are largely consistent across the three groups. We highlight where the results differ here. For the activation stage, $\beta_{\text{advice}}$ ($65\%$:$0.196$, $80\%$:$0.156$, $95\%$:$0.088$) does not have a significant effect on activation in the $95\%$ group; the advice is already perceived to be highly accurate and so it may not have as much weight in deciding whether to follow the advice. We also observe that $\beta_{\text{prior}}$ ($65\%$:$-0.003$, $80\%$:$0.223$, $95\%$:$-0.222$) is not consistent in sign across the three groups, suggesting there may be an effect of the perceived accuracy on the effect of the prior belief's utilization in decision making.

In the integration stage, we observe $\beta_{\text{advice}}$ ($65\%$:$-0.024$, $80\%$:$0.009$, $95\%$:$0.021$) is not consistent in sign across the three groups and is not significant in the $80\%$ group, suggesting the effect of advice confidence is not as clear as previously observed. This is not entirely surprising: note the coefficient magnitude is similar to those previously observed (Table~\ref{tbl:regression_coefficients}) and is more than $10x$ smaller than the magnitudes of other significant coefficients. This may suggest that while $\beta_{\text{advice}}$ has an effect on integration, it is not as important as other factors. 

\subsubsection{Location of prior belief question in survey}
In our experiments, we decided to include the prior belief survey question at the end of the survey, after all tasks had been completed. We intended to remove any potential biasing by introducing both human and AI advice into the experiment before we reveal that some participants have received human or AI advice. However, this results in a potential concern that participants' prior beliefs change over the course of the survey.

We check whether participants' prior beliefs changed in a systematic way by asking the prior belief survey question both before and after all tasks are completed for a small subset of our participants. We then check whether there is a statistically significant systematic change in prior belief value. We also fit the coefficients of our activation-integration model using both the start-of-survey responses and the end-of-survey responses. We observed no systematic change in prior belief. See Appendix~\ref{app:prior_survey} for more details.
\subsection{Real-world medical-AI task} \label{sec:biopsy_task}
\begin{figure}
\centering
\includegraphics[width=75mm]{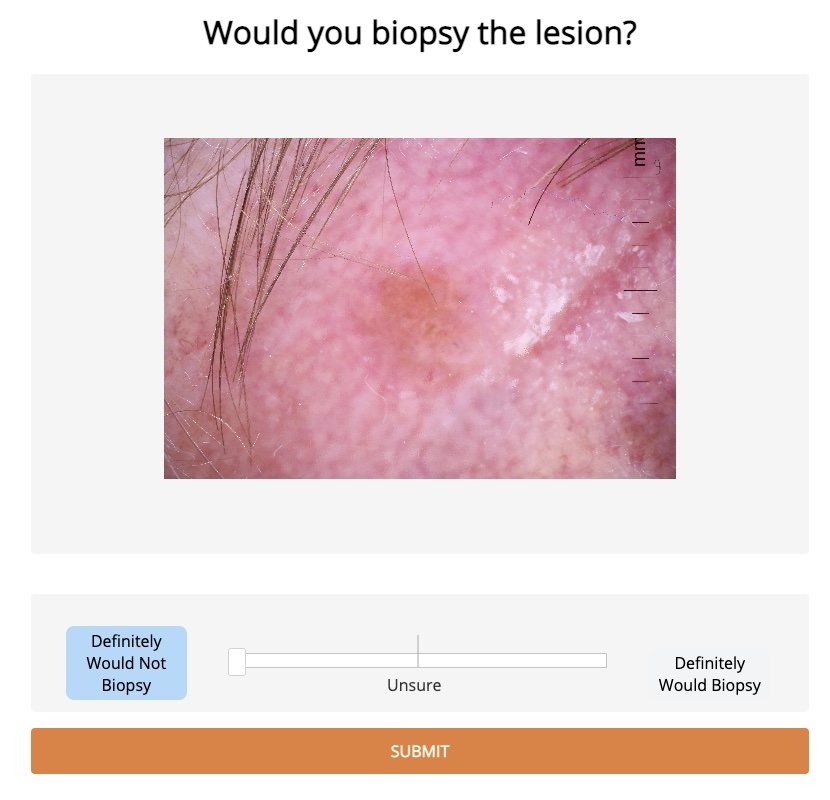}
\caption{Example task instance for the biopsy task.}
\label{fig:biopsy_task}
\end{figure}

While the 4 tasks described in Figure~\ref{fig:tasks} are ideal for crowdsourced data collection as they are accessible to the general public, it is unclear how well the observed results generalize to real-world tasks with experts (e.g., an AI medical assistant). To validate our activation-integration model on more realistic applications, we designed an additional task involving real medical data and recruited clinicians for our experiment. Additionally, we use a real AI model that we train to provide advice for the clinicians. We use the same experiment setup described in Figure~\ref{fig:study_design}.

The task is to determine whether a skin lesion should be biopsied or not based on a single image of the lesion (i.e., is the lesion likely malignant or benign); a sample task instance is shown in Figure~\ref{fig:biopsy_task}. The images used in the study were selected by a board-certified dermatologist from the International Skin Imaging Collaboration (ISIC) database. For the advice source, we trained a ResNet-18 model \cite{he2016deep} to predict whether a lesion is malignant or benign on the ISIC dataset \cite{rotemberg2021patient}. We then took the ResNet model output (after softmax) as our advice source. Across the data sample we show in our task, the model is well-calibrated with an expected calibration error of $0.068$.

Participants were dermatologists recruited by the authors through public outreach via email lists and social media. Most recruited dermatologists were practicing in the US (23); we also recruited dermatologists practicing in the UK (5), Korea (3), Japan (2), Canada (2), Brazil (2), and Poland (1). Unlike our other experiments, the recruited dermatologists were not given monetary compensation. Additionally, we did not measure socioeconomic scores or computer programming experience of the participants. Due to the more stringent recruitment parameters (the participants must be a practicing dermatologist), we were limited in the number of participants we could recruit. Demographic information on the recruited participants is provided in Table~\ref{tbl:biopsy_demographics}.

Performance was similar across the two conditions, with advice improving accuracy by a similar amount (Table~\ref{tbl:biopsy_results}), and only resulting in moderately higher average confidence in the correct answer. This concurs with prior work that found radiologists benefit similarly from human or AI advice when examining chest x-rays \cite{gaube2021ai}. The activation rate was also similar across human and AI advice. The initial and final accuracies are most comparable with the art task (for laypeople), where we observed a large difference in advice utilization across conditions ($4.7\%$ accuracy increase and $6.4\%$ activation increase for AI advice). These findings suggests that the dermatologists surveyed did not differentiate between human and AI advice. However, this may be a limitation of our study -- in practice, human advice generally comes from a known and trusted colleague rather than the unknown human source in our study.

We show our activation-integration model coefficients fit to the biopsy data in Table~\ref{tbl:biopsy_regression_coefficients}. As we could not collect the same demographic information as in previous experiments, we are limited in the number of factors we include in our model. However, we observe similar effects as previously described (Table~\ref{tbl:regression_coefficients}. The signs of $\beta_{\text{prior}}$, $\beta_{\text{response}}$, $\beta_{\text{advice}}$, and $\beta_{\text{consistent}}$ are consistent with our previous results. While $\beta_{\text{advice}}$ in the integration stage and $\beta_{\text{prior}}$ in both stages do not have a statistically significant effect, we can partially attribute this to the limited dataset size -- $<40\%$ of the data collected for our other experiments. 
We also note that $\beta_{\text{experience}}=-0.845$ is significant for activation -- suggesting that as a dermatologist becomes more experienced, they are less likely to heed advice. However, $\beta_{\text{experience}}$ does not have a significant effect in the integration setting, suggesting that experience serves to gate when advice is used but not how it is used, similar to the effect of prior belief discussed in the laypeople experiments.

These results are important as they indicate our activation-integration model largely holds up in a real-world setting. 

\begin{table}[h]
\centering
\caption{Participant demographics for Biopsy task.}
{\resizebox{\columnwidth}{!}{\begin{tabular}{lccccc}
\toprule
Dataset & \# of Participants & \# of Years Practicing & Education Level \\
\midrule
Biopsy & 37 &  8.6 +/- 8.6  & 8.0 \\
\bottomrule
\end{tabular}}}\label{tbl:biopsy_demographics}
\end{table}

\begin{table}[h]
\centering
\caption{Advice effect on Biopsy. ``Conf'' refers to the average confidence in the correct answer.}
{\resizebox{\columnwidth}{!}{\begin{tabular}{M{5em}M{5em}M{5em}M{5em}M{5em}}
\toprule
Advice source               & Baseline Accuracy (Conf) & Final Accuracy (Conf)  & Activation Rate  & Advice Accuracy \\
\midrule
AI    & 67.6\% (0.295)                 &  74.0\% (0.376)             & 29.7\%          & 83.3\%          \\\midrule
Human & 67.9\% (0.295)                 &  74.2\% (0.387)             & 29.8\%          & 83.3\%          \\
\bottomrule
\end{tabular}}}\label{tbl:biopsy_results}
\end{table}

\begin{table}[h]
\caption{Activation-integration model coefficients for Biopsy. Coefficients bolded if non-zero with confidence $>0.95$.}
{\resizebox{\columnwidth}{!}{\begin{tabular}{@{}M{10em}M{7em}M{7em}@{}}
\toprule
Covariates & Activation Coefficients & Response Change Coefficients \\
\midrule
Given AI Advice? ($\beta_{\text{ai}}$) & 0.143 +/- 0.602 & -0.002 +/- 0.063 \\
Prior Belief ($\beta_{\text{prior}}$) & -0.283 +/- 0.427 & 0.059 +/- 0.050 \\
Response 1 Confidence ($\beta_{\text{response}}$) & \textbf{-0.531 +/- 0.121} & \textbf{-0.201 +/- 0.026} \\
Advice Confidence ($\beta_{\text{advice}}$) & \textbf{0.260 +/- 0.111} & 0.024 +/- 0.024 \\
Response 1 and Advice Consistent? ($\beta_{\text{consistent}}$) & \textbf{-0.615 +/- 0.111} & \textbf{0.271 +/- 0.023} \\
Exposure to AI (in life) ($\beta_{\text{ai\_exposure}}$) & 1.185 +/- 1.269 & 0.147 +/- 0.140 \\
Years of Clinical Experience ($\beta_{\text{experience}}$) & \textbf{-0.845 +/- 0.411} & -0.056 +/- 0.051 \\
\bottomrule

\end{tabular}
}}\label{tbl:biopsy_regression_coefficients}
\end{table}
\section{Discussion and Future Work}
This work investigates how individuals incorporate external advice that comes from an AI algorithm. This is an important topic for the broader impact of AI and has been under-explored \cite{jussupow2020we,gaube2021ai}. We propose the \textit{activation-integration model} for advice utilization to explain our empirical findings and validate its utility across our diverse set of experimental settings. In the activation stage, a person decides whether to use the advice. In the integration stage, they decide how to incorporate the advice. We also find that the advice source affects whether the advice is acted on (activation), but now how it is integrated. This finding holds over a variety of settings with three different data modalities, across participants from the US, UK and Asia, and across perceived advice quality.

These results provide a better understanding of how humans incorporate advice from AI. Several exciting research directions lie ahead. For example, our work primarily focuses on common tasks---e.g. paintings, locations---that require no special expertise. This is a good starting point given that there has not been similar experiments before. An interesting question is to what extent our activation-integration model applies to settings where human experts incorporate external advise (e.g. a doctor is suggested a medical diagnoses by an AI or another doctor). Our initial results suggest the model holds; however, our data is limited and validating across other, similar tasks is important.
Another important question we can ask is how these results generalize across different types of advice. Here we focus on the most direct type of external feedback (i.e. a prediction with confidence score), because this is the most common type of output in machine learning. It would be interesting to explore more complex advice; e.g. where the AI provides an explanation together with its prediction.

\paragraph{Ethical Considerations}
We collect data from human subjects in our work through survey questions, where we asked crowd-sourced workers to complete simple tasks (image classification, sarcasm identification in text, and income level predictions from tabular data). We do not collect or release any personally identifiable data associated with the workers. Given the low risk nature of this data, IRB approval was not needed. For the medical task, we did obtain IRB approval. More information on the specifics of the data we collected including compensation and survey design are in Appendix~\ref{sec:app_compensation}.

This work and its potential extensions also raise some ethical questions. If it is possible to manipulate human utilization of AI advice either through the presentation of the algorithm or through the form of the advice given, there is risk that miscalibrated beliefs about the AI could harm performance. Note that this may not be done intentionally, but can still be harmful. So while understanding what factors influence human perception and utilization of AI advice is critical to designing systems where humans and AI work jointly, it must be done with care to ensure an AI agent is not over-trusted.


\bibliographystyle{ACM-Reference-Format}
\bibliography{references}


\begin{thebibliography}{32}


\ifx \showCODEN    \undefined \def \showCODEN     #1{\unskip}     \fi
\ifx \showDOI      \undefined \def \showDOI       #1{#1}\fi
\ifx \showISBNx    \undefined \def \showISBNx     #1{\unskip}     \fi
\ifx \showISBNxiii \undefined \def \showISBNxiii  #1{\unskip}     \fi
\ifx \showISSN     \undefined \def \showISSN      #1{\unskip}     \fi
\ifx \showLCCN     \undefined \def \showLCCN      #1{\unskip}     \fi
\ifx \shownote     \undefined \def \shownote      #1{#1}          \fi
\ifx \showarticletitle \undefined \def \showarticletitle #1{#1}   \fi
\ifx \showURL      \undefined \def \showURL       {\relax}        \fi
\providecommand\bibfield[2]{#2}
\providecommand\bibinfo[2]{#2}
\providecommand\natexlab[1]{#1}
\providecommand\showeprint[2][]{arXiv:#2}

\bibitem[\protect\citeauthoryear{Barr}{Barr}{2013}]%
        {barr2013random}
\bibfield{author}{\bibinfo{person}{Dale~J Barr}.}
  \bibinfo{year}{2013}\natexlab{}.
\newblock \showarticletitle{Random effects structure for testing interactions
  in linear mixed-effects models}.
\newblock \bibinfo{journal}{\emph{Frontiers in psychology}}
  \bibinfo{volume}{4} (\bibinfo{year}{2013}), \bibinfo{pages}{328}.
\newblock


\bibitem[\protect\citeauthoryear{Bates, M{\"a}chler, Bolker, and Walker}{Bates
  et~al\mbox{.}}{2014}]%
        {bates2014fitting}
\bibfield{author}{\bibinfo{person}{Douglas Bates}, \bibinfo{person}{Martin
  M{\"a}chler}, \bibinfo{person}{Ben Bolker}, {and} \bibinfo{person}{Steve
  Walker}.} \bibinfo{year}{2014}\natexlab{}.
\newblock \showarticletitle{Fitting linear mixed-effects models using lme4}.
\newblock \bibinfo{journal}{\emph{arXiv preprint arXiv:1406.5823}}
  (\bibinfo{year}{2014}).
\newblock


\bibitem[\protect\citeauthoryear{Chakrabarty and Biswas}{Chakrabarty and
  Biswas}{2018}]%
        {chakrabarty2018statistical}
\bibfield{author}{\bibinfo{person}{Navoneel Chakrabarty} {and}
  \bibinfo{person}{Sanket Biswas}.} \bibinfo{year}{2018}\natexlab{}.
\newblock \showarticletitle{A statistical approach to adult census income level
  prediction}. In \bibinfo{booktitle}{\emph{2018 International Conference on
  Advances in Computing, Communication Control and Networking (ICACCCN)}}.
  IEEE, \bibinfo{pages}{207--212}.
\newblock


\bibitem[\protect\citeauthoryear{De~Leeuw}{De~Leeuw}{2015}]%
        {de2015jspsych}
\bibfield{author}{\bibinfo{person}{Joshua~R De~Leeuw}.}
  \bibinfo{year}{2015}\natexlab{}.
\newblock \showarticletitle{jsPsych: A JavaScript library for creating
  behavioral experiments in a Web browser}.
\newblock \bibinfo{journal}{\emph{Behavior research methods}}
  \bibinfo{volume}{47}, \bibinfo{number}{1} (\bibinfo{year}{2015}),
  \bibinfo{pages}{1--12}.
\newblock


\bibitem[\protect\citeauthoryear{Dua and Graff}{Dua and Graff}{2017}]%
        {Dua:2019}
\bibfield{author}{\bibinfo{person}{Dheeru Dua} {and} \bibinfo{person}{Casey
  Graff}.} \bibinfo{year}{2017}\natexlab{}.
\newblock \bibinfo{title}{{UCI} Machine Learning Repository}.
\newblock
\newblock
\urldef\tempurl%
\url{https://archive.ics.uci.edu/ml/datasets/census+income}
\showURL{%
\tempurl}


\bibitem[\protect\citeauthoryear{Dzindolet, Pierce, Beck, and Dawe}{Dzindolet
  et~al\mbox{.}}{2002}]%
        {dzindolet2002perceived}
\bibfield{author}{\bibinfo{person}{Mary~T Dzindolet}, \bibinfo{person}{Linda~G
  Pierce}, \bibinfo{person}{Hall~P Beck}, {and} \bibinfo{person}{Lloyd~A
  Dawe}.} \bibinfo{year}{2002}\natexlab{}.
\newblock \showarticletitle{The perceived utility of human and automated aids
  in a visual detection task}.
\newblock \bibinfo{journal}{\emph{Human Factors}} \bibinfo{volume}{44},
  \bibinfo{number}{1} (\bibinfo{year}{2002}), \bibinfo{pages}{79--94}.
\newblock


\bibitem[\protect\citeauthoryear{Feldman, Aldana, and Stein}{Feldman
  et~al\mbox{.}}{2019}]%
        {feldman2019artificial}
\bibfield{author}{\bibinfo{person}{Robin~C Feldman}, \bibinfo{person}{Ehrik
  Aldana}, {and} \bibinfo{person}{Kara Stein}.}
  \bibinfo{year}{2019}\natexlab{}.
\newblock \showarticletitle{Artificial intelligence in the health care space:
  how we can trust what we cannot know}.
\newblock \bibinfo{journal}{\emph{Stan. L. \& Pol'y Rev.}}
  \bibinfo{volume}{30} (\bibinfo{year}{2019}), \bibinfo{pages}{399}.
\newblock


\bibitem[\protect\citeauthoryear{Fischler and Bolles}{Fischler and
  Bolles}{1981}]%
        {fischler1981random}
\bibfield{author}{\bibinfo{person}{Martin~A Fischler} {and}
  \bibinfo{person}{Robert~C Bolles}.} \bibinfo{year}{1981}\natexlab{}.
\newblock \showarticletitle{Random sample consensus: a paradigm for model
  fitting with applications to image analysis and automated cartography}.
\newblock \bibinfo{journal}{\emph{Commun. ACM}} \bibinfo{volume}{24},
  \bibinfo{number}{6} (\bibinfo{year}{1981}), \bibinfo{pages}{381--395}.
\newblock


\bibitem[\protect\citeauthoryear{Gaube, Suresh, Raue, Merritt, Berkowitz,
  Lermer, Coughlin, Guttag, Colak, and Ghassemi}{Gaube et~al\mbox{.}}{2021}]%
        {gaube2021ai}
\bibfield{author}{\bibinfo{person}{Susanne Gaube}, \bibinfo{person}{Harini
  Suresh}, \bibinfo{person}{Martina Raue}, \bibinfo{person}{Alexander Merritt},
  \bibinfo{person}{Seth~J Berkowitz}, \bibinfo{person}{Eva Lermer},
  \bibinfo{person}{Joseph~F Coughlin}, \bibinfo{person}{John~V Guttag},
  \bibinfo{person}{Errol Colak}, {and} \bibinfo{person}{Marzyeh Ghassemi}.}
  \bibinfo{year}{2021}\natexlab{}.
\newblock \showarticletitle{Do as AI say: susceptibility in deployment of
  clinical decision-aids}.
\newblock \bibinfo{journal}{\emph{NPJ digital medicine}} \bibinfo{volume}{4},
  \bibinfo{number}{1} (\bibinfo{year}{2021}), \bibinfo{pages}{1--8}.
\newblock


\bibitem[\protect\citeauthoryear{Gino and Moore}{Gino and Moore}{2007}]%
        {gino2007effects}
\bibfield{author}{\bibinfo{person}{Francesca Gino} {and} \bibinfo{person}{Don~A
  Moore}.} \bibinfo{year}{2007}\natexlab{}.
\newblock \showarticletitle{Effects of task difficulty on use of advice}.
\newblock \bibinfo{journal}{\emph{Journal of Behavioral Decision Making}}
  \bibinfo{volume}{20}, \bibinfo{number}{1} (\bibinfo{year}{2007}),
  \bibinfo{pages}{21--35}.
\newblock


\bibitem[\protect\citeauthoryear{Hagerty and Rubinov}{Hagerty and
  Rubinov}{2019}]%
        {hagerty2019global}
\bibfield{author}{\bibinfo{person}{Alexa Hagerty} {and} \bibinfo{person}{Igor
  Rubinov}.} \bibinfo{year}{2019}\natexlab{}.
\newblock \showarticletitle{Global AI ethics: a review of the social impacts
  and ethical implications of artificial intelligence}.
\newblock \bibinfo{journal}{\emph{arXiv preprint arXiv:1907.07892}}
  (\bibinfo{year}{2019}).
\newblock


\bibitem[\protect\citeauthoryear{Harvey and Fischer}{Harvey and
  Fischer}{1997}]%
        {harvey1997taking}
\bibfield{author}{\bibinfo{person}{Nigel Harvey} {and} \bibinfo{person}{Ilan
  Fischer}.} \bibinfo{year}{1997}\natexlab{}.
\newblock \showarticletitle{Taking advice: Accepting help, improving judgment,
  and sharing responsibility}.
\newblock \bibinfo{journal}{\emph{Organizational behavior and human decision
  processes}} \bibinfo{volume}{70}, \bibinfo{number}{2} (\bibinfo{year}{1997}),
  \bibinfo{pages}{117--133}.
\newblock


\bibitem[\protect\citeauthoryear{Harvey, Harries, and Fischer}{Harvey
  et~al\mbox{.}}{2000}]%
        {harvey2000using}
\bibfield{author}{\bibinfo{person}{Nigel Harvey}, \bibinfo{person}{Clare
  Harries}, {and} \bibinfo{person}{Ilan Fischer}.}
  \bibinfo{year}{2000}\natexlab{}.
\newblock \showarticletitle{Using advice and assessing its quality}.
\newblock \bibinfo{journal}{\emph{Organizational behavior and human decision
  processes}} \bibinfo{volume}{81}, \bibinfo{number}{2} (\bibinfo{year}{2000}),
  \bibinfo{pages}{252--273}.
\newblock


\bibitem[\protect\citeauthoryear{He, Zhang, Ren, and Sun}{He
  et~al\mbox{.}}{2016}]%
        {he2016deep}
\bibfield{author}{\bibinfo{person}{Kaiming He}, \bibinfo{person}{Xiangyu
  Zhang}, \bibinfo{person}{Shaoqing Ren}, {and} \bibinfo{person}{Jian Sun}.}
  \bibinfo{year}{2016}\natexlab{}.
\newblock \showarticletitle{Deep residual learning for image recognition}. In
  \bibinfo{booktitle}{\emph{Proceedings of the IEEE conference on computer
  vision and pattern recognition}}. \bibinfo{publisher}{IEEE},
  \bibinfo{pages}{770--778}.
\newblock


\bibitem[\protect\citeauthoryear{Jussupow, Benbasat, and Heinzl}{Jussupow
  et~al\mbox{.}}{2020}]%
        {jussupow2020we}
\bibfield{author}{\bibinfo{person}{Ekaterina Jussupow}, \bibinfo{person}{Izak
  Benbasat}, {and} \bibinfo{person}{Armin Heinzl}.}
  \bibinfo{year}{2020}\natexlab{}.
\newblock \showarticletitle{Why are we averse towards Algorithms? A
  comprehensive literature Review on Algorithm aversion.}. In
  \bibinfo{booktitle}{\emph{ECIS}}.
\newblock


\bibitem[\protect\citeauthoryear{Khodak, Saunshi, and Vodrahalli}{Khodak
  et~al\mbox{.}}{2017}]%
        {khodak2017large}
\bibfield{author}{\bibinfo{person}{Mikhail Khodak}, \bibinfo{person}{Nikunj
  Saunshi}, {and} \bibinfo{person}{Kiran Vodrahalli}.}
  \bibinfo{year}{2017}\natexlab{}.
\newblock \showarticletitle{A large self-annotated corpus for sarcasm}.
\newblock \bibinfo{journal}{\emph{arXiv preprint arXiv:1704.05579}}
  (\bibinfo{year}{2017}).
\newblock


\bibitem[\protect\citeauthoryear{Lee}{Lee}{2018}]%
        {lee2018understanding}
\bibfield{author}{\bibinfo{person}{Min~Kyung Lee}.}
  \bibinfo{year}{2018}\natexlab{}.
\newblock \showarticletitle{Understanding perception of algorithmic decisions:
  Fairness, trust, and emotion in response to algorithmic management}.
\newblock \bibinfo{journal}{\emph{Big Data \& Society}} \bibinfo{volume}{5},
  \bibinfo{number}{1} (\bibinfo{year}{2018}),
  \bibinfo{pages}{2053951718756684}.
\newblock


\bibitem[\protect\citeauthoryear{Madhavan and Wiegmann}{Madhavan and
  Wiegmann}{2007}]%
        {madhavan2007effects}
\bibfield{author}{\bibinfo{person}{Poornima Madhavan} {and}
  \bibinfo{person}{Douglas~A Wiegmann}.} \bibinfo{year}{2007}\natexlab{}.
\newblock \showarticletitle{Effects of information source, pedigree, and
  reliability on operator interaction with decision support systems}.
\newblock \bibinfo{journal}{\emph{Human Factors}} \bibinfo{volume}{49},
  \bibinfo{number}{5} (\bibinfo{year}{2007}), \bibinfo{pages}{773--785}.
\newblock


\bibitem[\protect\citeauthoryear{Mesbah, Tauchert, and Buxmann}{Mesbah
  et~al\mbox{.}}{2021}]%
        {mesbah2021whose}
\bibfield{author}{\bibinfo{person}{Neda Mesbah}, \bibinfo{person}{Christoph
  Tauchert}, {and} \bibinfo{person}{Peter Buxmann}.}
  \bibinfo{year}{2021}\natexlab{}.
\newblock \showarticletitle{Whose Advice Counts More--Man or Machine? An
  Experimental Investigation of AI-based Advice Utilization}. In
  \bibinfo{booktitle}{\emph{Proceedings of the 54th Hawaii International
  Conference on System Sciences}}. \bibinfo{publisher}{ScholarSpace},
  \bibinfo{pages}{4083}.
\newblock


\bibitem[\protect\citeauthoryear{Miller}{Miller}{2019}]%
        {miller2019explanation}
\bibfield{author}{\bibinfo{person}{Tim Miller}.}
  \bibinfo{year}{2019}\natexlab{}.
\newblock \showarticletitle{Explanation in artificial intelligence: Insights
  from the social sciences}.
\newblock \bibinfo{journal}{\emph{Artificial intelligence}}
  \bibinfo{volume}{267} (\bibinfo{year}{2019}), \bibinfo{pages}{1--38}.
\newblock


\bibitem[\protect\citeauthoryear{{\"O}nkal, Goodwin, Thomson, G{\"o}n{\"u}l,
  and Pollock}{{\"O}nkal et~al\mbox{.}}{2009}]%
        {onkal2009relative}
\bibfield{author}{\bibinfo{person}{Dilek {\"O}nkal}, \bibinfo{person}{Paul
  Goodwin}, \bibinfo{person}{Mary Thomson}, \bibinfo{person}{Sinan
  G{\"o}n{\"u}l}, {and} \bibinfo{person}{Andrew Pollock}.}
  \bibinfo{year}{2009}\natexlab{}.
\newblock \showarticletitle{The relative influence of advice from human experts
  and statistical methods on forecast adjustments}.
\newblock \bibinfo{journal}{\emph{Journal of Behavioral Decision Making}}
  \bibinfo{volume}{22}, \bibinfo{number}{4} (\bibinfo{year}{2009}),
  \bibinfo{pages}{390--409}.
\newblock


\bibitem[\protect\citeauthoryear{Prahl and Van~Swol}{Prahl and
  Van~Swol}{2017}]%
        {prahl2017understanding}
\bibfield{author}{\bibinfo{person}{Andrew Prahl} {and} \bibinfo{person}{Lyn
  Van~Swol}.} \bibinfo{year}{2017}\natexlab{}.
\newblock \showarticletitle{Understanding algorithm aversion: When is advice
  from automation discounted?}
\newblock \bibinfo{journal}{\emph{Journal of Forecasting}}
  \bibinfo{volume}{36}, \bibinfo{number}{6} (\bibinfo{year}{2017}),
  \bibinfo{pages}{691--702}.
\newblock


\bibitem[\protect\citeauthoryear{Prolific}{Prolific}{2022}]%
        {prolific}
Prolific \bibinfo{year}{2022}\natexlab{}.
\newblock \bibinfo{title}{Prolific Academic}.
\newblock \bibinfo{howpublished}{\url{https://www.prolific.co}}.
\newblock


\bibitem[\protect\citeauthoryear{Ribeiro, Singh, and Guestrin}{Ribeiro
  et~al\mbox{.}}{2016}]%
        {ribeiro2016should}
\bibfield{author}{\bibinfo{person}{Marco~Tulio Ribeiro},
  \bibinfo{person}{Sameer Singh}, {and} \bibinfo{person}{Carlos Guestrin}.}
  \bibinfo{year}{2016}\natexlab{}.
\newblock \showarticletitle{"Why should i trust you?" Explaining the
  predictions of any classifier}. In \bibinfo{booktitle}{\emph{Proceedings of
  the 22nd ACM SIGKDD international conference on knowledge discovery and data
  mining}}. \bibinfo{publisher}{ACM}, \bibinfo{pages}{1135--1144}.
\newblock


\bibitem[\protect\citeauthoryear{Rotemberg, Kurtansky, Betz-Stablein, Caffery,
  Chousakos, Codella, Combalia, Dusza, Guitera, Gutman,
  et~al\mbox{.}}{Rotemberg et~al\mbox{.}}{2021}]%
        {rotemberg2021patient}
\bibfield{author}{\bibinfo{person}{Veronica Rotemberg},
  \bibinfo{person}{Nicholas Kurtansky}, \bibinfo{person}{Brigid Betz-Stablein},
  \bibinfo{person}{Liam Caffery}, \bibinfo{person}{Emmanouil Chousakos},
  \bibinfo{person}{Noel Codella}, \bibinfo{person}{Marc Combalia},
  \bibinfo{person}{Stephen Dusza}, \bibinfo{person}{Pascale Guitera},
  \bibinfo{person}{David Gutman}, {et~al\mbox{.}}}
  \bibinfo{year}{2021}\natexlab{}.
\newblock \showarticletitle{A patient-centric dataset of images and metadata
  for identifying melanomas using clinical context}.
\newblock \bibinfo{journal}{\emph{Scientific data}} \bibinfo{volume}{8},
  \bibinfo{number}{1} (\bibinfo{year}{2021}), \bibinfo{pages}{1--8}.
\newblock


\bibitem[\protect\citeauthoryear{Sah, Moore, and MacCoun}{Sah
  et~al\mbox{.}}{2013}]%
        {sah2013cheap}
\bibfield{author}{\bibinfo{person}{Sunita Sah}, \bibinfo{person}{Don~A Moore},
  {and} \bibinfo{person}{Robert~J MacCoun}.} \bibinfo{year}{2013}\natexlab{}.
\newblock \showarticletitle{Cheap talk and credibility: The consequences of
  confidence and accuracy on advisor credibility and persuasiveness}.
\newblock \bibinfo{journal}{\emph{Organizational Behavior and Human Decision
  Processes}} \bibinfo{volume}{121}, \bibinfo{number}{2}
  (\bibinfo{year}{2013}), \bibinfo{pages}{246--255}.
\newblock


\bibitem[\protect\citeauthoryear{Schultze, Rakotoarisoa, and
  Schulz-Hardt}{Schultze et~al\mbox{.}}{2015}]%
        {schultze2015effects}
\bibfield{author}{\bibinfo{person}{Thomas Schultze},
  \bibinfo{person}{Anne-Fernandine Rakotoarisoa}, {and} \bibinfo{person}{Stefan
  Schulz-Hardt}.} \bibinfo{year}{2015}\natexlab{}.
\newblock \showarticletitle{Effects of distance between initial estimates and
  advice on advice utilization.}
\newblock \bibinfo{journal}{\emph{Judgment \& Decision Making}}
  \bibinfo{volume}{10}, \bibinfo{number}{2} (\bibinfo{year}{2015}).
\newblock


\bibitem[\protect\citeauthoryear{Tauchert, Mesbah, et~al\mbox{.}}{Tauchert
  et~al\mbox{.}}{2019}]%
        {tauchert2019following}
\bibfield{author}{\bibinfo{person}{Christoph Tauchert}, \bibinfo{person}{Neda
  Mesbah}, {et~al\mbox{.}}} \bibinfo{year}{2019}\natexlab{}.
\newblock \showarticletitle{Following the Robot? Investigating Users'
  Utilization of Advice from Robo-Advisors.}. In
  \bibinfo{booktitle}{\emph{ICIS}}. \bibinfo{publisher}{AIS}.
\newblock


\bibitem[\protect\citeauthoryear{Van~Swol, Paik, and Prahl}{Van~Swol
  et~al\mbox{.}}{2018}]%
        {van2018advice}
\bibfield{author}{\bibinfo{person}{Lyn~M Van~Swol},
  \bibinfo{person}{Jihyun~Esther Paik}, {and} \bibinfo{person}{Andrew Prahl}.}
  \bibinfo{year}{2018}\natexlab{}.
\newblock \showarticletitle{Advice recipients: The psychology of advice
  utilization.}
\newblock  (\bibinfo{year}{2018}).
\newblock


\bibitem[\protect\citeauthoryear{West and Praturu}{West and Praturu}{2019}]%
        {census2019}
\bibfield{author}{\bibinfo{person}{Alex West} {and} \bibinfo{person}{Anusha
  Praturu}.} \bibinfo{year}{2019}\natexlab{}.
\newblock \bibinfo{title}{Enhancing the Census Income Prediction Dataset}.
\newblock
  \bibinfo{howpublished}{\url{https://people.ischool.berkeley.edu/~alexwest/w210_census_income_html/}}.
\newblock
\newblock
\shownote{Accessed: 2021-05-15.}


\bibitem[\protect\citeauthoryear{Xie, Chen, Kao, Gao, and Chen}{Xie
  et~al\mbox{.}}{2020}]%
        {xie2020chexplain}
\bibfield{author}{\bibinfo{person}{Yao Xie}, \bibinfo{person}{Melody Chen},
  \bibinfo{person}{David Kao}, \bibinfo{person}{Ge Gao}, {and}
  \bibinfo{person}{Xiang'Anthony' Chen}.} \bibinfo{year}{2020}\natexlab{}.
\newblock \showarticletitle{CheXplain: Enabling Physicians to Explore and
  Understand Data-Driven, AI-Enabled Medical Imaging Analysis}. In
  \bibinfo{booktitle}{\emph{Proceedings of the 2020 CHI Conference on Human
  Factors in Computing Systems}}. \bibinfo{publisher}{ACM},
  \bibinfo{pages}{1--13}.
\newblock


\bibitem[\protect\citeauthoryear{Yaniv and Foster}{Yaniv and Foster}{1997}]%
        {yaniv1997precision}
\bibfield{author}{\bibinfo{person}{Ilan Yaniv} {and} \bibinfo{person}{Dean~P
  Foster}.} \bibinfo{year}{1997}\natexlab{}.
\newblock \showarticletitle{Precision and accuracy of judgmental estimation}.
\newblock \bibinfo{journal}{\emph{Journal of behavioral decision making}}
  \bibinfo{volume}{10}, \bibinfo{number}{1} (\bibinfo{year}{1997}),
  \bibinfo{pages}{21--32}.
\newblock


\end{thebibliography}

\appendix

\section{Appendix}\label{app:overview}
We provide more data about participants' prior beliefs about the effectiveness of AI in Section~\ref{sec:prior}. Section~\ref{sec:UKAsia} provides more details about our UK and Asia experiments which further validated our findings. Section~\ref{sec:survey} provides more information about our survey design. Screenshots of our experiment survey are provided as Supplementary Files; see Section~\ref{sec:survey_screens}. Our study followed standard practices for studying human evaluations of external advice \cite{van2018advice}. All participants consented to join the study and we protect their privacy by only using de-identified data in our analysis.

\section{Distribution of prior}\label{sec:prior}

\begin{figure*}[h]
\begin{tabular}{cc}
  \includegraphics[width=65mm]{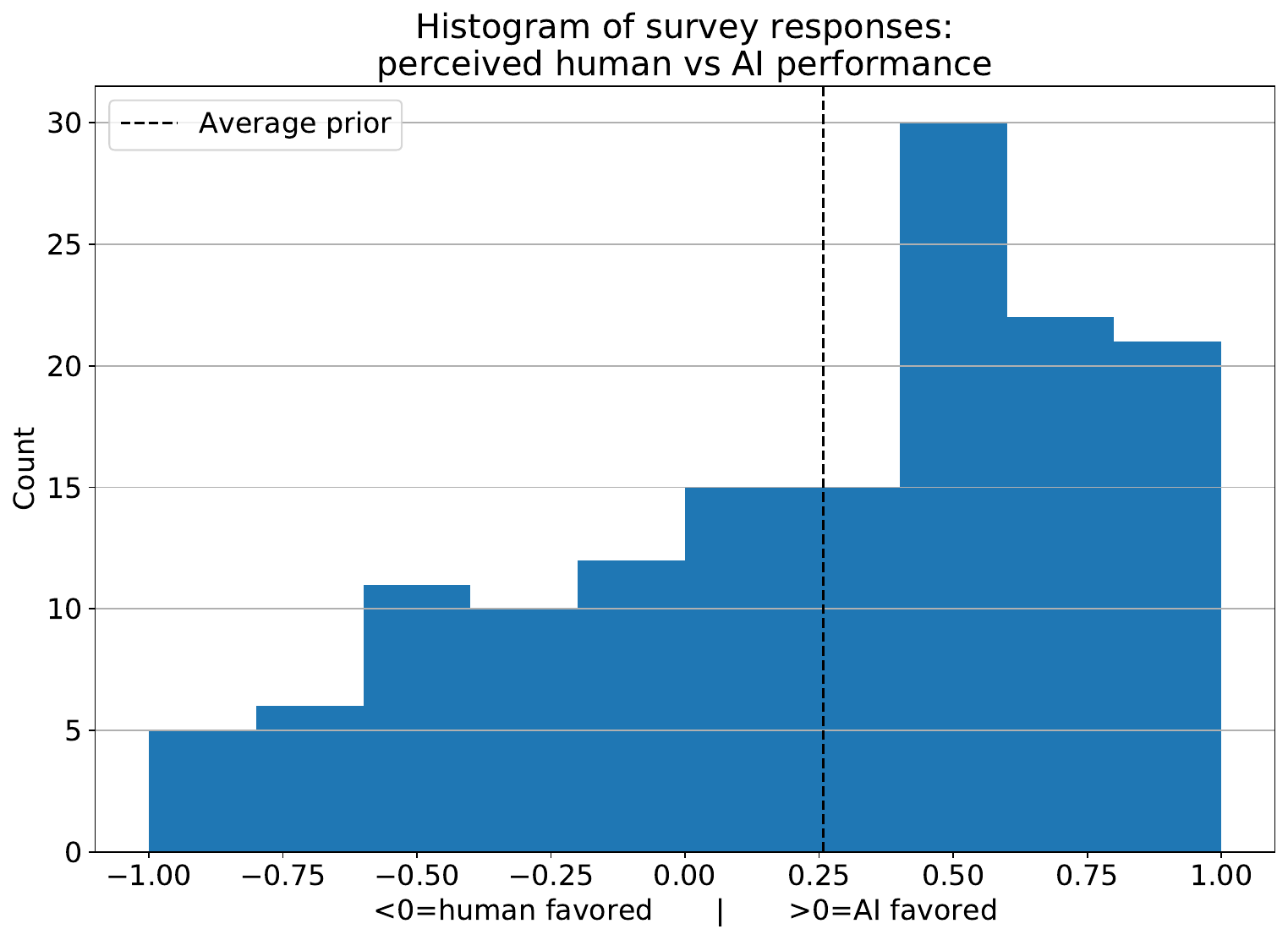} &  \includegraphics[width=65mm]{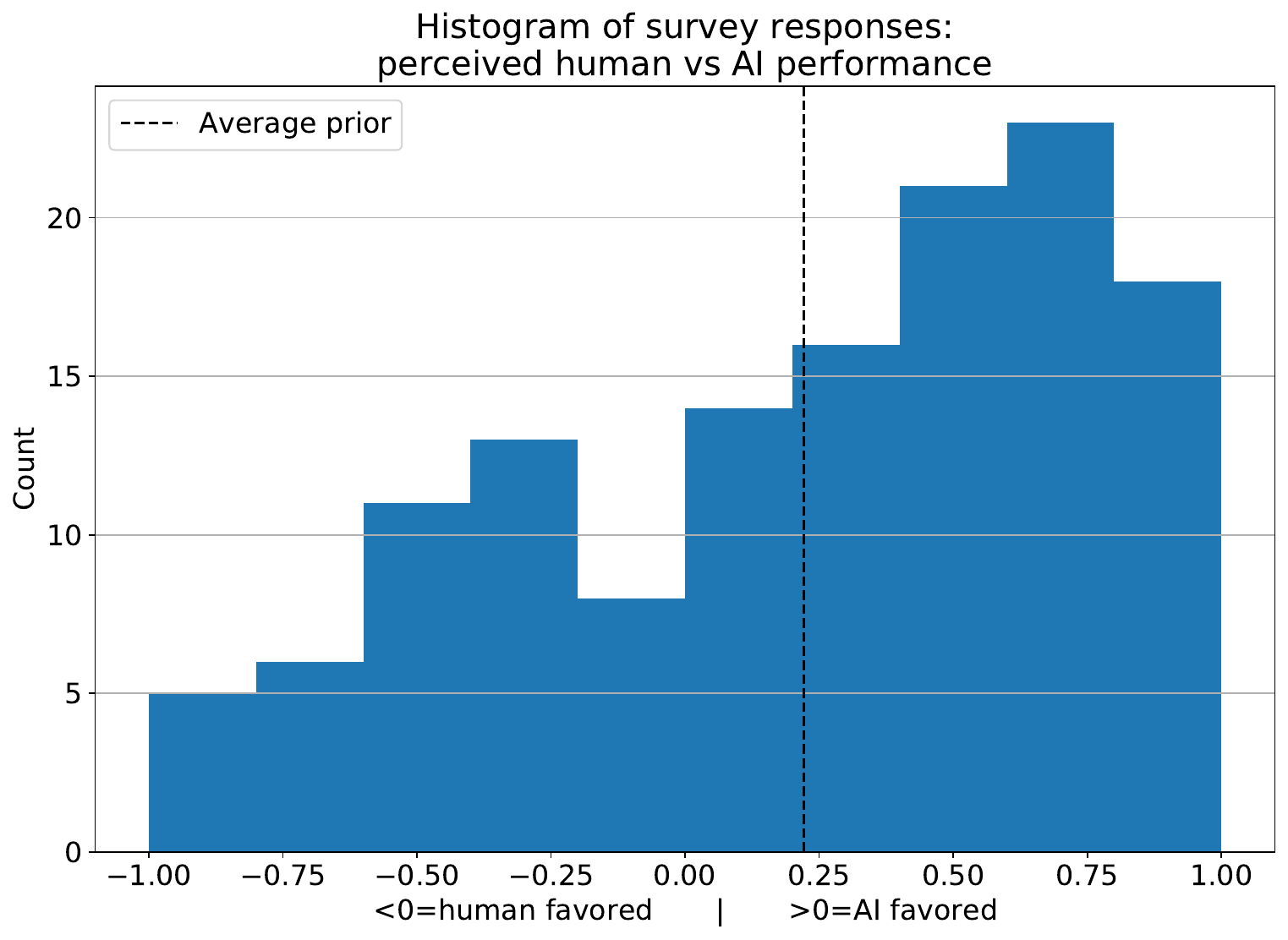} \\ (a) Art & (b) Cities \\[6pt]
 \includegraphics[width=65mm]{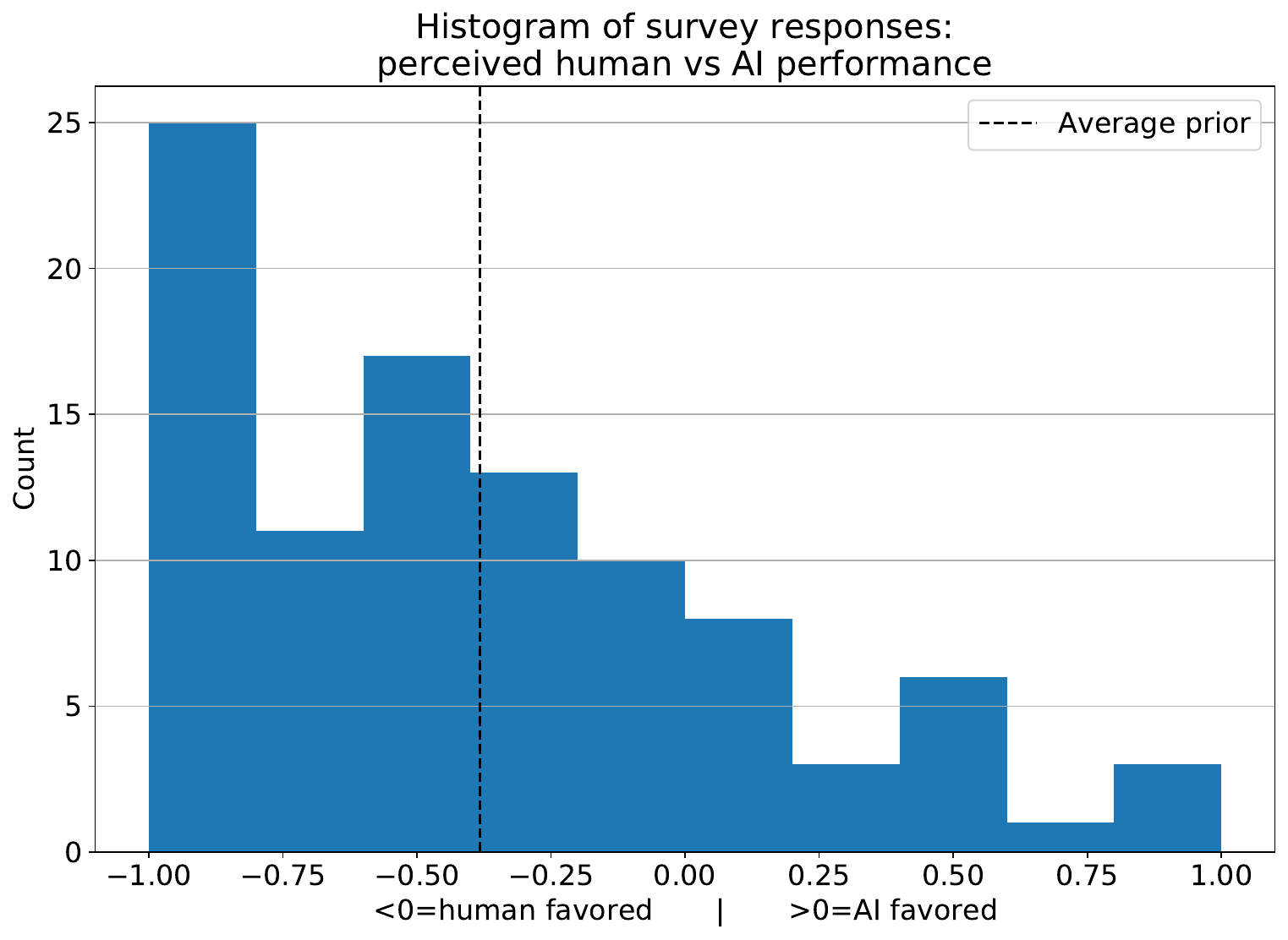} & \includegraphics[width=65mm]{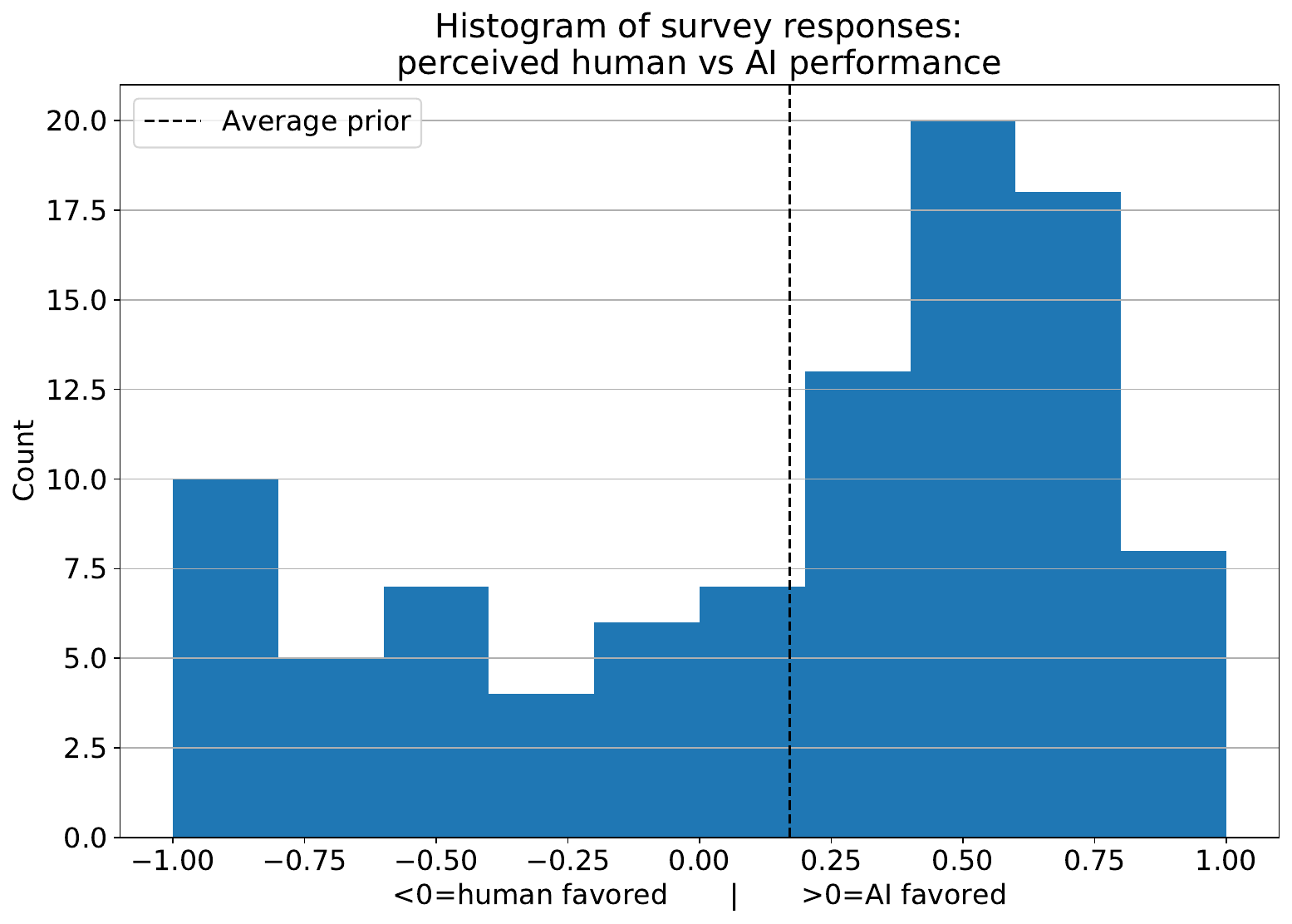} \\ (c) Sarcasm & (d) Census \\[6pt]
\end{tabular}
\caption{Histogram of the prior belief across each of our 4 datasets (for US participants). Prior belief is computed from a survey question (see Section~\ref{sec:prior_survey}) and normalized to $[-1,1]$. $<0$ indicates human-favored; $>0$ indicates AI-favored. 
The histogram includes both participants who received human advice and participants who received AI advice. 
The dotted vertical line is the average prior across participants.}
\label{fig:prior_distribution}
\end{figure*}

We show the distributions of prior belief across our four datasets in Figure~\ref{fig:prior_distribution}. For the art, cities, and census datasets, the prior belief favors the AI advice; for the sarcasm dataset, the prior belief favors the human advice. Note that for all datasets there is a high variance, and there are always participants who strongly believe either the AI or human advice is better regardless of the average belief.

\section{UK and Asia validation experiments} \label{sec:UKAsia}

\subsection{Dataset Overview}
We show an overview of the datasets used for experiments in the UK and Asia in Table~\ref{tbl:datasets_summary_global}. Comparing with Table~\ref{tbl:datasets_summary}, note that the baseline accuracies in the UK are similar to the US, while the baseline accuracies in Asia are lower in both datasets The prior belief across both datasets in Asia and the UK is in the same direction as in the US (AI or human favored). In the UK, the prior belief is stronger (e.g., higher percentage of people favor AI for the art dataset and human for the sarcasm dataset); in Asia, it is only stronger for the art dataset.

\begin{table*}[h]
\caption{Summary of the datasets we use in our experiments in the UK and Asia. Follows same format as Table~\ref{tbl:datasets_participant_summary}}
{\resizebox{2\columnwidth}{!}{\begin{tabular}{@{}lM{7em}M{10em}M{7em}M{7em}r@{}}
\toprule
Dataset & Data Type & Description & \# of Tasks & Baseline accuracy & Prior belief \\
\midrule
Art (UK) & Image & identify the art period & 32 & 65.5\% +/- 16.6\% & 74.4\% (AI) \\
Sarcasm (UK) & Text & identify sarcasm & 32 & 70.6\% +/- 20.8\% & 83.0\% (human) \\
\midrule
Art (Asia) & Image & identify the art period & 32 & 62.2\% +/- 17.0\% & 76.3\% (AI) \\
Sarcasm (Asia) & Text & identify sarcasm & 32 & 62.1\% +/- 21.9\% & 78.2\% (human) \\
\bottomrule
\end{tabular}}
}\label{tbl:datasets_summary_global}
\end{table*}

\subsection{Demographics}
Demographic information for the UK and Asia experiments is shown in Table~\ref{tbl:datasets_participant_summary_global}. All values are similar to the US datasets with two slight differences: (1) the average age of the Sarcasm dataset in the UK is close to 10 years higher, and (2) the socioeconomic score and education level in Asia are both about 0.5 higher than the US data.

Participants from Asia came from a diverse set of countries. At least 10 participants across both art and sarcasm datasets are Indian, Chinese, Vietnamese, Filipino, Pakistani, Indonesian, or Korean.

\begin{table*}[h]
\caption{Summary of the participant demographics from the 4 datasets we use in our experiments. See Section~\ref{sec:demographic_details} for information on the socioeconomic score and education level.}
{\resizebox{2\columnwidth}{!}{\begin{tabular}{@{}lM{7em}M{7em}M{7em}M{7em}M{7em}@{}}
\toprule
Dataset & \# of Participants & Gender (Percent Male / Female) & Age & Socioeconomic Score & Education Level \\
\midrule
Art Dataset (UK) & 86 & 45.3\% / 54.7\% & 32.2 +/- 12.3 & 5.2 +/- 1.6 & 5.1 +/- 1.4 \\
Sarcasm Dataset (UK) & 88 & 55.7\% / 44.3\% & 41.4 +/- 15.4 & 4.9 +/- 1.8 & 5.1 +/- 1.6 \\
Art Dataset (Asia) & 97 & 56.7\% / 43.3\% & 27.4 +/- 6.8 & 5.8 +/- 1.6 & 5.9 +/- 1.1 \\
Sarcasm Dataset (Asia) & 101 & 48.5\% / 51.5\% & 29.0 +/- 7.3 & 5.8 +/- 1.4 & 6.1 +/- 1.3 \\
\bottomrule
\end{tabular}}
}\label{tbl:datasets_participant_summary_global}
\end{table*}

\subsection{Activation-integration model}
Here we include the coefficient values referenced in Section~\ref{sec:regression_global}. In Table~\ref{tbl:regression_coefficients_global_logistic}, we show the coefficients for the activation model trained using US data, UK data, and Asia data. As was noted earlier, the magnitude and sign of the $\beta_{\text{prior}},$  $\beta_{\text{response}},$  $\beta_{\text{advice}},$ and  $\beta_{\text{consistent}}$ are largely the same across all three regions. The single exception is the prior effect in Asia is smaller, and subsequently we do not have enough data to conclude significance.

In Table~\ref{tbl:regression_coefficients_global_linear}, we show the coefficients for the integration model across all three regions. $\beta_{\text{response}},$  $\beta_{\text{advice}},$ and  $\beta_{\text{consistent}}$ have identical sign and similar magnitude across all three regions. The prior effect is not significant in any of the regions, supporting our two stage model.

\begin{table*}[h]
\caption{Coefficients for activation stage mixed effects logistic regression model. Comparison across different geographical regions.}
{\resizebox{2\columnwidth}{!}{\begin{tabular}{@{}lM{7em}M{7em}M{7em}@{}}
\toprule
Covariates & Coefficients (US) & Coefficients (UK) & Coefficients (Asia) \\
\midrule
Given AI Advice? ($\beta_{\text{ai}}$) & 0.162 +/- 0.134 & -0.253 +/- 0.218 & 0.389 +/- 0.207 \\
Prior Belief ($\beta_{\text{prior}}$) & \textbf{0.244 +/- 0.059} & \textbf{0.217 +/- 0.107} & 0.101 +/- 0.100 \\
Response 1 Confidence ($\beta_{\text{response}}$) & \textbf{-0.584 +/- 0.025} & \textbf{-0.491 +/- 0.037} & \textbf{-0.355 +/- 0.038} \\
Advice Confidence ($\beta_{\text{advice}}$) & \textbf{0.218 +/- 0.024} & \textbf{0.146 +/- 0.036} & \textbf{0.143 +/- 0.033} \\
Response 1 and Advice Consistent? ($\beta_{\text{consistent}}$) & \textbf{-1.555 +/- 0.051} & \textbf{-1.157 +/- 0.076} & \textbf{-1.492 +/- 0.073} \\
Age ($\beta_{\text{age}}$) & -0.089 +/- 0.072 & -0.074 +/- 0.084 & 0.110 +/- 0.192 \\
Gender ($\beta_{\text{gender}}$) & -0.046 +/- 0.137 & 0.201 +/- 0.229 & 0.157 +/- 0.224 \\
Education ($\beta_{\text{education}}$) & 0.043 +/- 0.076 & 0.027 +/- 0.106 & -0.118 +/- 0.133 \\
Socioeconomic Status ($\beta_{\text{socioeconomic}}$) & \textbf{0.139 +/- 0.070} & 0.013 +/- 0.109 & -0.010 +/- 0.109 \\
Has Programming Experience? ($\beta_{\text{programming}}$) & 0.233 +/- 0.147 & -0.209 +/- 0.256 & 0.031 +/- 0.217 \\
\bottomrule

\end{tabular}
}}\label{tbl:regression_coefficients_global_logistic}
\end{table*}

\begin{table*}[h]
\caption{Coefficients for integration stage mixed effects linear regression model. Comparison across different geographical regions.}
{\resizebox{2\columnwidth}{!}{\begin{tabular}{@{}lM{7em}M{7em}M{7em}@{}}
\toprule
Covariates & Coefficients (US) & Coefficients (UK) & Coefficients (Asia) \\
\midrule
Given AI Advice? ($\beta_{\text{ai}}$) & -0.023 +/- 0.016 & 0.008 +/- 0.024 & -0.000 +/- 0.027 \\
Prior Belief ($\beta_{\text{prior}}$) & 0.001 +/- 0.008 & -0.023 +/- 0.012 & 0.005 +/- 0.013 \\
Response 1 Confidence ($\beta_{\text{response}}$) & \textbf{-0.212 +/- 0.004} & \textbf{-0.199 +/- 0.007} & \textbf{-0.226 +/- 0.007} \\
Advice Confidence ($\beta_{\text{advice}}$) & \textbf{0.025 +/- 0.004} & \textbf{0.033 +/- 0.007} & \textbf{0.016 +/- 0.006} \\
Response 1 and Advice Consistent? ($\beta_{\text{consistent}}$) & \textbf{0.661 +/- 0.008} & \textbf{0.589 +/- 0.014} & \textbf{0.674 +/- 0.011} \\
Age ($\beta_{\text{age}}$) & \textbf{-0.026 +/- 0.008} & -0.013 +/- 0.009 & -0.014 +/- 0.025 \\
Gender ($\beta_{\text{gender}}$) & 0.010 +/- 0.016 & 0.027 +/- 0.025 & -0.031 +/- 0.029 \\
Education ($\beta_{\text{education}}$) & \textbf{0.021 +/- 0.009} & -0.016 +/- 0.012 & 0.015 +/- 0.017 \\
Socioeconomic Status ($\beta_{\text{socioeconomic}}$) & 0.007 +/- 0.008 & -0.020 +/- 0.012 & \textbf{-0.028 +/- 0.014} \\
Has Programming Experience? ($\beta_{\text{programming}}$) & 0.014 +/- 0.017 & 0.044 +/- 0.028 & -0.007 +/- 0.028 \\
\bottomrule

\end{tabular}
}}\label{tbl:regression_coefficients_global_linear}
\end{table*}

\section{Perceived accuracy validation experiments} \label{sec:perceived_acc_val}

\subsection{Dataset Overview}
We show an overview of the datasets used for experiments on the perceived accuracy variation in Table~\ref{tbl:datasets_summary_acc}. We use the art dataset (Figure~\ref{fig:tasks} a); the $80\%$ accurate advice data is the same data as used in previous sections. Note the only difference between the three datasets is the information given to the participants about the accuracy of the advice; the actual advice used is identical across all three datasets. Demographic values are observed to be similar across the three datasets, as expected.

\begin{table*}[h]
\caption{Summary of the datasets we use in our experiments varying the perceived accuracy. Follows same format as Table~\ref{tbl:datasets_participant_summary}.}
{\resizebox{2\columnwidth}{!}{\begin{tabular}{@{}lM{7em}M{7em}M{7em}M{7em}M{7em}r@{}}
\toprule
Dataset & \# of Participants & Gender (Percent Male / Female) & Age & Socioeconomic Score & Education Level \\
\midrule
Art (65\% accurate advice) & 103 & 42.7\% / 57.3\% & 34.1 +/- 12.8 & 5.0 +/- 1.5 & 5.7 +/- 1.2 \\
Art (80\% accurate advice) & 147 & 47.6\% / 52.4\% & 33.0 +/- 12.3 & 5.2 +/- 1.5 & 5.4 +/- 1.3 \\
Art (95\% accurate advice) & 104 & 39.4 \% / 60.6\% & 29.0 +/- 9.8 & 4.9 +/- 1.7 & 5.3 +/- 1.2 \\
\bottomrule
\end{tabular}}
}\label{tbl:datasets_summary_acc}
\end{table*}

\subsection{Activation-integration model}
Here we include the coefficient values referenced in Section~\ref{sec:perceived_acc}. In Table~\ref{tbl:regression_coefficients_acc_logistic}, we show the coefficients for the activation model trained using the three perceived accuracy bins. As was noted earlier, the magnitude and sign of the $\beta_{\text{prior}},$  $\beta_{\text{response}},$  $\beta_{\text{advice}},$ and  $\beta_{\text{consistent}}$ are largely the same across all three groups.

In Table~\ref{tbl:regression_coefficients_acc_linear}, we show the coefficients for the integration model across all three groups. $\beta_{\text{response}},$  $\beta_{\text{advice}},$ and  $\beta_{\text{consistent}}$ have identical sign and similar magnitude across all three groups. The prior effect is not significant in any of the regions, supporting our two stage model.

\begin{table*}[h]
\caption{Coefficients for activation stage mixed effects logistic regression model. Comparison across different perceived advice accuracies.}
{\resizebox{2\columnwidth}{!}{\begin{tabular}{@{}lM{7em}M{7em}M{7em}@{}}
\toprule
Covariates & Coefficients (65\%) & Coefficients (80\%) & Coefficients (95\%) \\
\midrule
Given AI Advice? ($\beta_{\text{ai}}$) & -0.006 +/- 0.297 & 0.167 +/- 0.278 & 0.582 +/- 0.339 \\
Prior Belief ($\beta_{\text{prior}}$) & -0.003 +/- 0.176 & 0.223 +/- 0.151 & -0.222 +/- 0.156 \\
Response 1 Confidence ($\beta_{\text{response}}$) & \textbf{-0.729 +/- 0.058} & \textbf{-0.679 +/- 0.050} & \textbf{-0.665 +/- 0.059} \\
Advice Confidence ($\beta_{\text{advice}}$) & \textbf{0.196 +/- 0.052} & \textbf{0.156 +/- 0.042} & 0.088 +/- 0.052 \\
Response 1 and Advice Consistent? ($\beta_{\text{consistent}}$) & \textbf{-1.616 +/- 0.103} & \textbf{-2.066 +/- 0.093} & \textbf{-2.004 +/- 0.114} \\
Age ($\beta_{\text{age}}$) & -0.155 +/- 0.134 & 0.123 +/- 0.118 & 0.006 +/- 0.173 \\
Gender ($\beta_{\text{gender}}$) & 0.173 +/- 0.306 & -0.001 +/- 0.264 & -0.348 +/- 0.311 \\
Education ($\beta_{\text{education}}$) & -0.141 +/- 0.174 & -0.132 +/- 0.136 & -0.180 +/- 0.170 \\
Socioeconomic Status ($\beta_{\text{socioeconomic}}$) & 0.323 +/- 0.167 & 0.220 +/- 0.137 & \textbf{0.387 +/- 0.151} \\
Has Programming Experience? ($\beta_{\text{programming}}$) & -0.198 +/- 0.342 & 0.251 +/- 0.283 & -0.495 +/- 0.346 \\
\bottomrule

\end{tabular}
}}\label{tbl:regression_coefficients_acc_logistic}
\end{table*}

\begin{table*}[h]
\caption{Coefficients for integration stage mixed effects linear regression model. Comparison across different perceived advice accuracies.}
{\resizebox{2\columnwidth}{!}{\begin{tabular}{@{}lccr@{}}

\toprule
Covariates & Coefficients (65\%) & Coefficients (80\%) & Coefficients (95\%) \\
\midrule
Given AI Advice? ($\beta_{\text{ai}}$) & 0.046 +/- 0.045 & \textbf{-0.082 +/- 0.039} & -0.019 +/- 0.035 \\
Prior Belief ($\beta_{\text{prior}}$) & -0.038 +/- 0.028 & 0.017 +/- 0.021 & -0.026 +/- 0.016 \\
Response 1 Confidence ($\beta_{\text{response}}$) & \textbf{-0.186 +/- 0.012} & \textbf{-0.237 +/- 0.009} & \textbf{-0.259 +/- 0.010} \\
Advice Confidence ($\beta_{\text{advice}}$) & \textbf{-0.024 +/- 0.010} & 0.009 +/- 0.008 & \textbf{0.021 +/- 0.009} \\
Response 1 and Advice Consistent? ($\beta_{\text{consistent}}$) & \textbf{0.599 +/- 0.019} & \textbf{0.681 +/- 0.014} & \textbf{0.677 +/- 0.017} \\
Age ($\beta_{\text{age}}$) & -0.011 +/- 0.021 & -0.028 +/- 0.016 & -0.011 +/- 0.018 \\
Gender ($\beta_{\text{gender}}$) & 0.011 +/- 0.047 & -0.010 +/- 0.037 & 0.029 +/- 0.032 \\
Education ($\beta_{\text{education}}$) & -0.018 +/- 0.026 & 0.004 +/- 0.019 & 0.027 +/- 0.018 \\
Socioeconomic Status ($\beta_{\text{socioeconomic}}$) & 0.035 +/- 0.026 & 0.030 +/- 0.019 & -0.012 +/- 0.016 \\
Has Programming Experience? ($\beta_{\text{programming}}$) & -0.001 +/- 0.052 & 0.031 +/- 0.039 & -0.002 +/- 0.036 \\
\bottomrule

\end{tabular}
}}\label{tbl:regression_coefficients_acc_linear}
\end{table*}

\section{Survey Details}\label{sec:survey}

\subsection{Demographic Information Collection} \label{sec:demographic_details}
\begin{figure*}[h]
\centerline{\includegraphics[width=13cm]{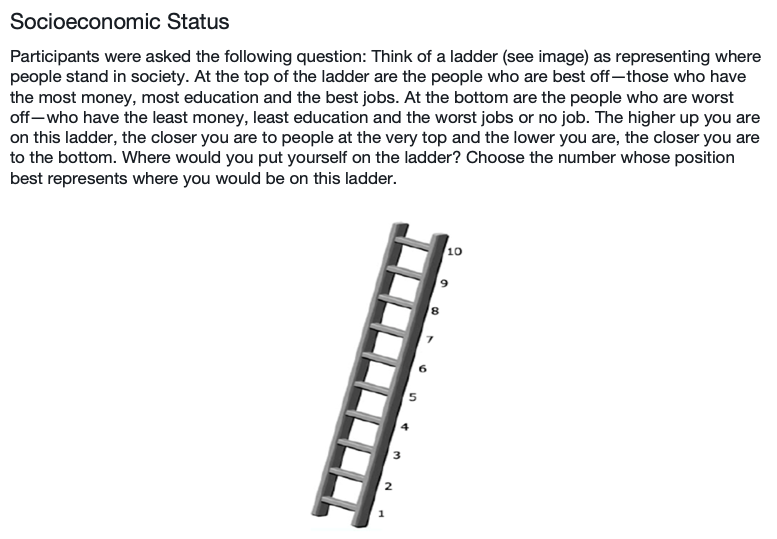}}
\caption{Question participants are asked by Prolific \cite{prolific} to obtain participant's socioeconomic status.}
\label{fig:socioeconomic_status}
\end{figure*}

We conducted our experiments using Prolific \cite{prolific}. Prolific provides access to various demographic information that we used in our study. These include age, gender, education level, socioeconomic status, and whether the participant has programming experience. Education level is defined on a scale from 1 to 8. The interpretation of the education level is given below:
\begin{enumerate}
    \item[1 --]Don't know / not applicable
    \item[2 --]No formal qualifications
    \item[3 --]Secondary education (e.g. GED/GCSE)
    \item[4 --]High school diploma/A-levels
    \item[5 --]Technical/community college
    \item[6 --]Undergraduate degree (BA/BSc/other)
    \item[7 --]Graduate degree (MA/MSc/MPhil/other)
    \item[8 --]Doctorate degree (PhD/other)
\end{enumerate}
Socioeconomic status is defined on a scale from 1 to 10. The question asked to participants to determine the socioeconomic status is included in Figure~\ref{fig:socioeconomic_status}.

\subsection{Survey Question About Prior} \label{app:prior_survey}
Here we add some additional details on the experiments we ran to check whether completing the task has a systematic effect on prior belief response. As described in the main paper, we asked the survey question to measure prior belief after all task instances have been completed in the majority of our experiments. This was done to avoid any potential biasing in the participants -- prior to completing the tasks, the participants are not aware that there are two types of advice they could have received. However, a concern is that completing the task instances will bias a participant's response. 

To check whether the location of the survey affected people's responses, we ran additional experiments on the Art and Sarcasm datasets where we asked the prior belief survey question both prior to showing participants any tasks and after all tasks are completed.

\begin{figure*}[ht]
\begin{tabular}{cc}
  \includegraphics[width=75mm]{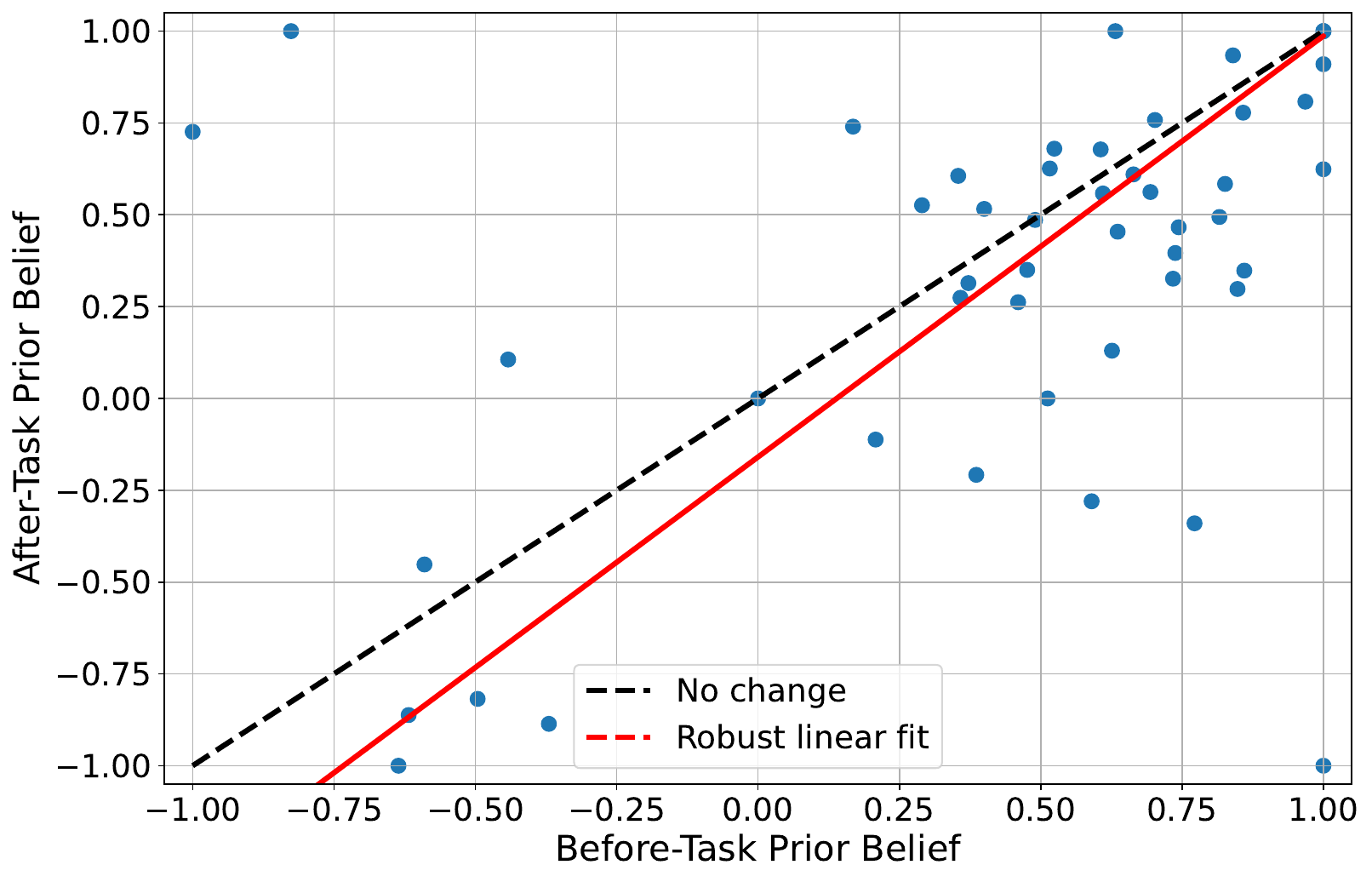} &  \includegraphics[width=75mm]{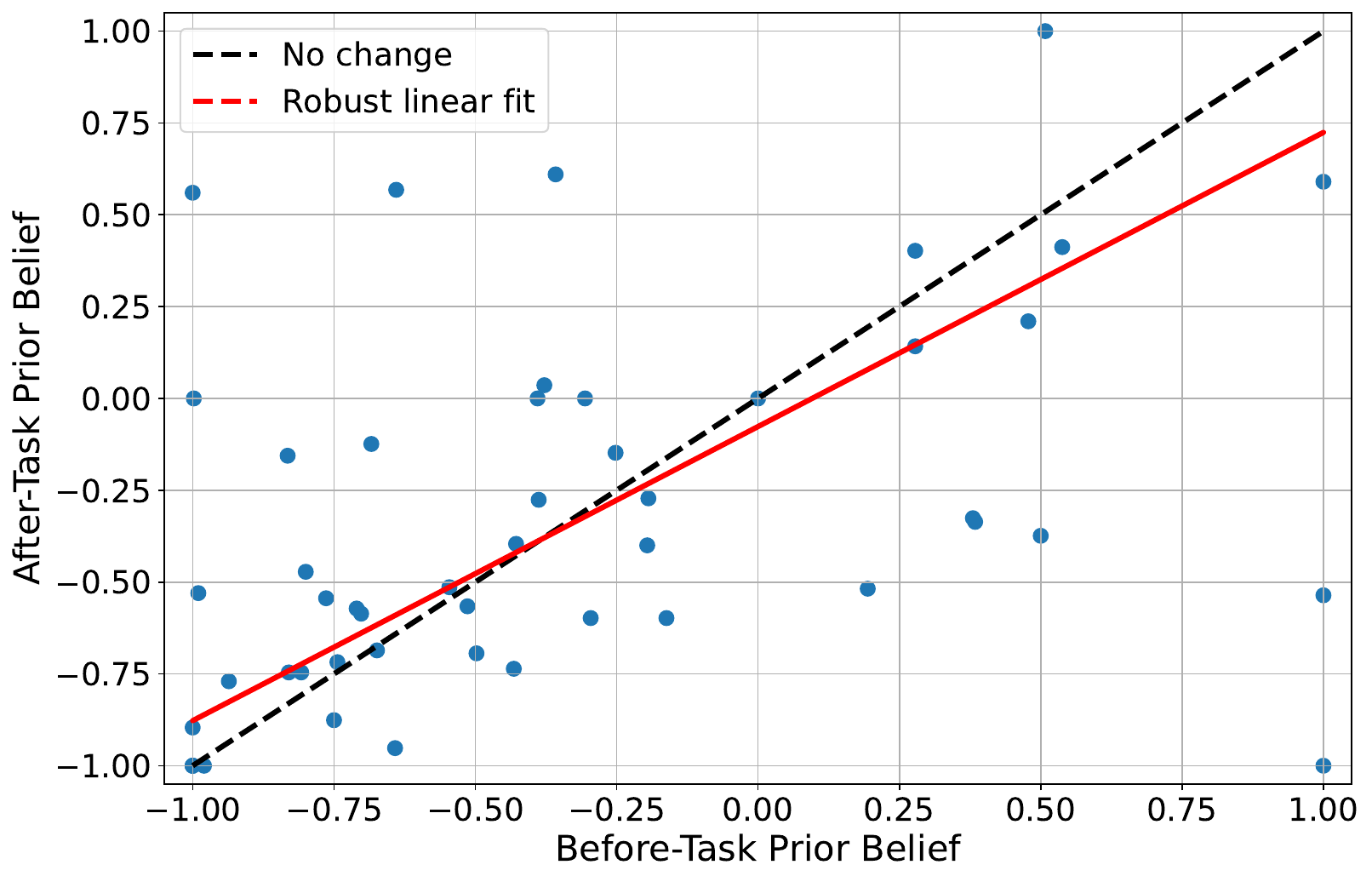} \\ (a) Art & (b) Sarcasm \\[6pt]
\end{tabular}
\caption{Scatter plot of prior belief survey response before and after tasks for (a) Art and (b) Sarcasm datasets. The dotted black line shows the expected linear fit if location of the survey does not affect prior belief response; the red line shows a robust linear fit of the data.}
\label{fig:prior_survey_robust}
\end{figure*}

In Figure~\ref{fig:prior_survey_robust}, we plot the prior belief survey response before and after the participant has completed all task instances. We plot a robust linear fit of the data computed using RANSAC \cite{fischler1981random} (in red) as well as the expected fit if prior belief response is not affected by completing the task instances. The robust fit in both the Art and Sarcasm datasets indicate that the location of prior belief survey does not have a systematic affect on prior belief response.

We further confirmed this conclusion by running our mixed effects analysis using the prior belief responses from the start of the survey. We found that our main findings hold---prior belief has a significant effect for the activation stage, but not for the integration stage.

\subsection{Participant Compensation}\label{sec:app_compensation}
Participants were compensated at a rate of $\$10.00$ per hour as per the Prolific recommended rates. The survey was estimated to take 10 minutes based on several trial runs by the authors. Participants were compensated with this assumption.

Participants were also informed that they could receive up to a 30\% bonus. This bonus was calculated as 
$$\text{bonus} = \begin{cases}
0 & \text{if } S < 0.3 \\
S*0.3 & \text{otherwise}
\end{cases},$$
where $S$ is the average performance of the participant across all tasks, both before and after receiving advice. Performance for a single task is computed as 
$$ sign(\text{correct response}) \cdot \text{response}_1 $$ 
where $-1 \leq \text{response}_1 \leq 1$. Note that this performance metric penalizes incorrect responses.

The total cost of running all of our experiments (including the participants we used to calibrate the advice) was around $\$2,500$.

\newpage
\section{Survey Screenshots} \label{sec:survey_screens}

We designed our study using standard methods in the psychology of advice utilization literature \cite{van2018advice}. The study was implemented for web deployment using jsPsych \cite{de2015jspsych} and a simple Python-based web server. In the ``survey-art-dataset'' directory of the Supplementary material, we include a set of PDF documents showing screenshots of our survey for the Art dataset for a participant who received human advice. Each “page” in the directory corresponds to a separate web page. Filenames are numbered in the order a participant encounters them. Clicking “continue” / “submit” brings the participant to the next web page.

A brief description of the survey screenshots follows. Any content referring to the art data or AI advice was substituted with appropriate content for each dataset and human advice respectively. 

\begin{enumerate}
    \item[\textbf{1:}] Participant enters a unique ID assigned to them through Prolific \cite{prolific}, the crowdsource platform we use to recruit participants. 
    \item[\textbf{2-4:}] Instructions specific to the task and advice source participants will receive. Note that this page is seen by participants as a single, continuous web page.
    \item[\textbf{5:}] Additional information on the advice source. (For participants receiving human advice, the text is changed accordingly.)
    \item[\textbf{6:}] Information on bonus payment.
    \item[\textbf{7:}] Manipulation check. Participants who got the wrong answer were sent back to Page 2.
    \item[\textbf{8:}] Example task: recording Response 1. The ``initial response'' block in Figure~\ref{fig:study_design} shows the bottom half of this slide (for the Cities dataset). Figure~\ref{fig:tasks} shows example images of this slide for each of our four datasets.
    \item[\textbf{9:}] Example task: recording Response 2. The ``Human advice'' block in Figure~\ref{fig:study_design} shows the bottom half of this slide (for the Cities dataset).
    \item[\textbf{10:}] After completing all tasks, Participants are shown this screen.
    \item[\textbf{11-15:}] Additional survey questions. The response on slide 12 is used for the belief prior.
    \item[\textbf{16:}] Check for errors in survey. Primarily used when developing the experiment to ensure the survey was bug-free. 
    \item[\textbf{17:}] Debrief slide. 
    \item[\textbf{18:}] Bonus payment and survey submission slide.
\end{enumerate}

\clearpage


\section*{Supplementary material}

Please enter your **Prolific ID:**

Prolific ID: En

Continue

# Instructions:

In this study, you will be labeling 32 different paintings / croppings of a painting based on what art period the painting is from. Do **NOT** lookup try to search for paintings from the art periods while completing this survey.
**Bonus payment is available!** You will be given an additional monetary reward of up to 30% the base payment based on how well you perform.
Follow the figures below for instructions.

Once you have finished reading, **click the "Continue" button at the bottom.**

**Step 1. (1)** You will see an image appear in the center of the screen. **(2)** A slider bar will appear below it. **(3)** Click a location on the slider bar, based on which art period you think the painting is from. The location you select should correspond

to your confidence in predicting the art period the painting is from. For example, we clicked a location close to "Definitely Renaissance (~1300-1600)" because this is the Mona Lisa, a painting from the Renaissance period. Once you have selected a location, click "SUBMIT".

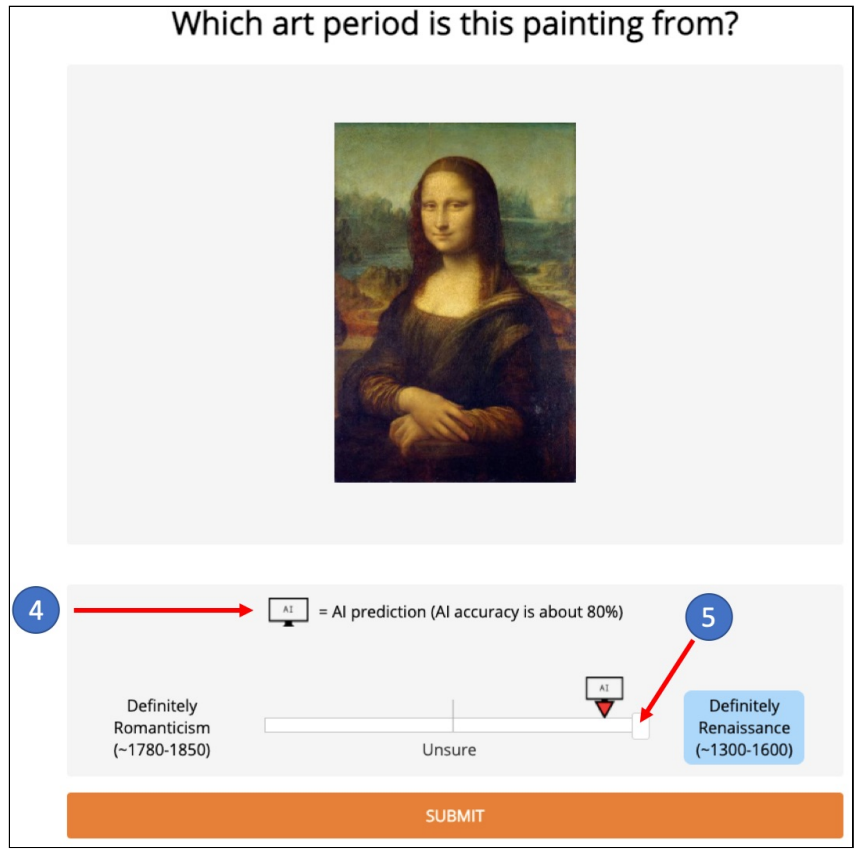

**Step 2.** After clicking "SUBMIT", a **(4)** computer icon will appear. This computer icon represents the prediction of an artificial intelligence (AI) algorithm on this same task; in this example, the AI thought the painting was from the Renaissance period. **(5)** As before, select a location on the slider bar using this new information. You must click on the slider before submitting, even if you want to keep your selection. Once you have selected a location, click "SUBMIT".

19

You will then be shown a new image. Repeat the process from **Steps 1 and 2** until there are no more images left to label. After labeling the images, you will have a brief survey to complete and you will be done.

Continue

# Instructions -- Advice from artificial intelligence (AI) model:

After labeling an image on your own, we will show you the output of an **AI** algorithm trained to identify the art period a painting was created in.

The **AI** has an average accuracy of around **80%**.

Continue

# Instructions -- Bonus payment:

You will receive bonus pay based on your performance! We will compute your average accuracy to determine your bonus.

Simply put, if you put a high confidence and get the correct answer, you will get a larger bonus. But if you put a high confidence and get the answer wrong, your bonus is reduced. If you are unsure about an answer, **it is best for you to adjust the slider to reflect your confidence**.

Continue

What is the average accuracy of the AI algorithm?*

60% ○    70% ○    80% ○    90% ○    100% ○

Continue

# Which art period is this painting from?

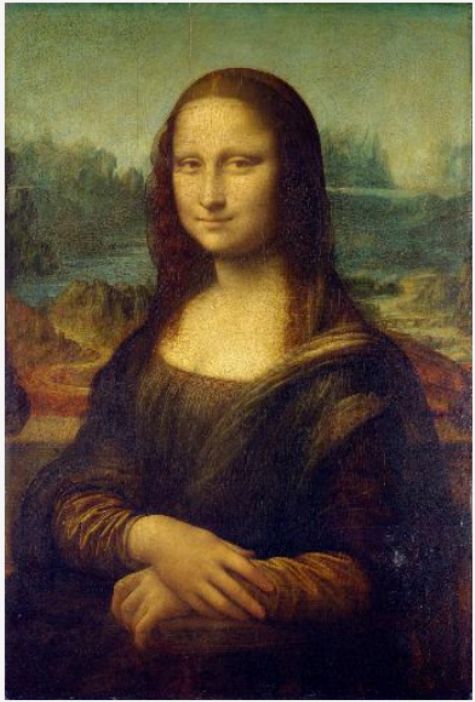

| Definitely Romanticism (~1780-1850) | 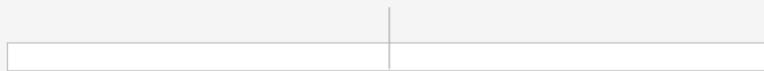 Unsure | Definitely Renaissance (~1300-1600) |

SUBMIT

# Which art period is this painting from?

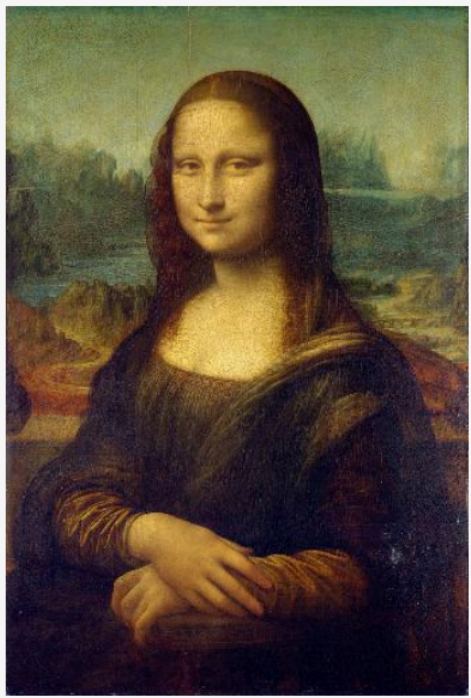

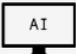
= AI prediction (AI accuracy is about 80%)

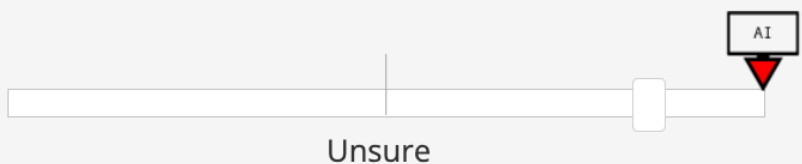

Definitely Romanticism (~1780-1850) — Unsure — Definitely Renaissance (~1300-1600)

SUBMIT

# Which art period is this painting from?

You are finished labeling images.

Press the 'Enter'/'Return' key

on your keyboard to continue.

| Definitely [ArtPeriod1] | Unsure | Definitely [ArtPeriod2] |

**SUBMIT**

# Exit survey:

Please answer the following questions about the labeling task. There is no right/wrong answer here, and you should answer based on your personal experience.

**1. How accurate were the human labelers at determining the correct art period?**

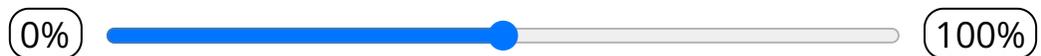

Continue

# Exit survey:

Please answer the following questions about the labeling task. There is no right/wrong answer here, and you should answer based on your personal experience.

**2. How much did the advice from the human labelers help you?**

(Not at all) —————————————————— (Very much)

Continue

# Exit survey:

Please answer the following questions about the labeling task. There is no right/wrong answer here, and you should answer based on your personal experience.

**3. How much do you trust the human labelers?**

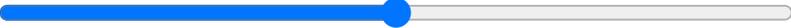

Continue

# Exit survey:

Please answer the following questions about the labeling task. There is no right/wrong answer here, and you should answer based on your personal experience.

**4. Do you think an artificial intelligence (AI) algorithm or the average person (without help) can do better on this task?**

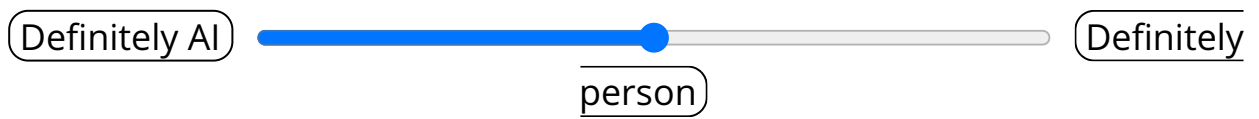

Continue

# Exit survey:

Please answer the following questions about the labeling task. There is no right/wrong answer here, and you should answer based on your personal experience.

**5. How often do you use artificial intelligence (AI) systems to aid you in your everyday life and/or at work?**

Never ———————————○——————— Very often

Continue

# Were there any errors you encountered while going through the survey?

[ ]

Continue

# DEBRIEF:

In this survey, you were tasked with labeling paintings by the art period they were created in. You were additionally given some advice to help you complete this task. People completing this survey were split into 2 groups and either given advice from **a previous group of human labelers** (what you received) or an AI model. However, both groups were actually given the same advice (the advice is based on data from a previous group of human labelers, and **not** an AI model in both cases).

The purpose of this experiment was to investigate whether people would respond differently to advice coming from humans vs. advice coming from an AI. To ensure that any difference in response was only based on perception and trust in AI algorithms vs. humans, we used the same advice for both groups.

This type of work is important for a few reasons. As AI technology has rapidly matured in the last decade, one of the biggest barriers to adoption is the limited trust people have in models they don't understand. This is particularly important in medicine, where doctors are often unwilling to trust an AI -- unlike when talking with people, an AI model's prediction cannot easily be verified or justified by asking the model why it made its prediction.

Continue

You received a bonus of: $0.51

Please click the following link to complete the survey:
https://app.prolific.co/submissions/complete?cc=9145FD8C



\end{document}